\RequirePackage{fix-cm}
\documentclass[twocolumn]{svjour3}          
\smartqed  
\usepackage{graphicx}
\usepackage{multirow}
\usepackage{times}
\usepackage{epsfig}
\usepackage{graphicx}
\usepackage{amsmath}
\usepackage{amssymb}
\usepackage{subfigure}
\usepackage{graphicx}
\usepackage{amsmath,amssymb} 
\usepackage{multirow}
\usepackage{algorithmic}
\usepackage{algorithm}
\usepackage{bbm}
\usepackage{hyperref}
\usepackage{color}
\usepackage{epstopdf}
\usepackage[T1]{fontenc}
\usepackage[utf8]{inputenc}
\usepackage[font=small,labelfont=bf,tableposition=top]{caption}
\usepackage{booktabs}
\usepackage{threeparttable}

\DeclareMathAlphabet{\mathpzc}{OT1}{pzc}{m}{it} %
\DeclareMathOperator*{\argmax}{argmax}

\newcommand{\trans}[1]{{#1}^{\ensuremath{\mathsf{T}}}} 

\newenvironment{packed_enumerate}{
	\begin{enumerate}
		\setlength{\itemsep}{1pt}
		\setlength{\parskip}{0pt}
		\setlength{\parsep}{0pt}
	}{\end{enumerate}}

\journalname{Submission to International Journal of Computer Vision}

\newcommand{\etal} {\textit{et al.}}
\newcommand{\ie} {\textit{i.e.}}
\newcommand{\eg} {\textit{e.g.}}
\newcommand{\x} {\textbf{x}}

\newcommand{\g} {\textbf{g}}

\newcommand{\Y} {\textbf{Y}}
\newcommand{\w} {\textbf{w}}
\newcommand{\W} {\textbf{W}}
\newcommand{\X} {\textbf{X}}
\newcommand{\I} {\textbf{I}}
\newcommand{\K} {\textbf{K}}

\newcommand{\kk} {\textbf{k}}

\newcommand{\Q} {\textbf{Q}}

\begin{document}

\title{From Facial Expression Recognition to Interpersonal Relation Prediction
}


\author{Zhanpeng Zhang         \and
        Ping Luo \and Chen Change Loy \and Xiaoou Tang 
}


\institute{
Zhanpeng Zhang \at
SenseTime Group Limited \\
\email{zhangzhanpeng@sensetime.com}
\and
Ping Luo \at
Department of Information Engineering, The Chinese University of Hong Kong \\
\email{pluo@ie.cuhk.edu.hk}
\and 
Chen Change Loy (Corresponding Author) \at
Department of Information Engineering, The Chinese University of Hong Kong \\
\email{ccloy@ie.cuhk.edu.hk}
\and
Xiaoou Tang \at
Department of Information Engineering, The Chinese University of Hong Kong \\
\email{xtang@ie.cuhk.edu.hk}
}

\date{Received: date / Accepted: date}

\maketitle

\begin{abstract}

Interpersonal relation defines the association, \eg, warm, friendliness, and dominance, between two or more people.
We investigate if such fine-grained and high-level relation traits can be
characterized and quantified from face images in the wild. 
We address this challenging problem by first studying a deep network architecture for robust recognition of facial expressions. Unlike existing models that typically learn from facial expression labels alone, we devise an effective multitask network that is capable of learning from rich auxiliary attributes such as gender, age, and head pose, beyond just facial expression data. While conventional supervised training requires datasets with complete labels (\eg, all samples must be labeled with gender, age, and expression), we show that this requirement can be relaxed via a novel attribute propagation method. The approach further allows us to leverage the inherent correspondences between heterogeneous attribute sources despite the disparate distributions of different datasets.
With the network we demonstrate state-of-the-art results on existing facial expression recognition benchmarks.
To predict inter-personal relation, we use the expression recognition network as branches for a Siamese model.
Extensive experiments show that our model is capable of mining mutual context of faces for accurate fine-grained interpersonal prediction. 

\end{abstract}

\section{Introduction}
\label{sec:intro}

Facial expression recognition is an actively researched topic in computer vision~\cite{tian2011facial}. 
Existing pipelines typically recognize single-person expressions and assign them into discrete prototypical classes, namely anger, disgust, fear, happy, sad, surprise, and neutral.  
Inspired by extensive psychological studies~\cite{Girard:2014,gottman2001facial,hess2000influence,Knutsonfd}, in this work we wish to investigate the interesting problem of characterizing and quantifying interpersonal relation traits from human face images beyond just expressions. 

\begin{figure}[t]
	\centering
	\includegraphics[width=\linewidth]{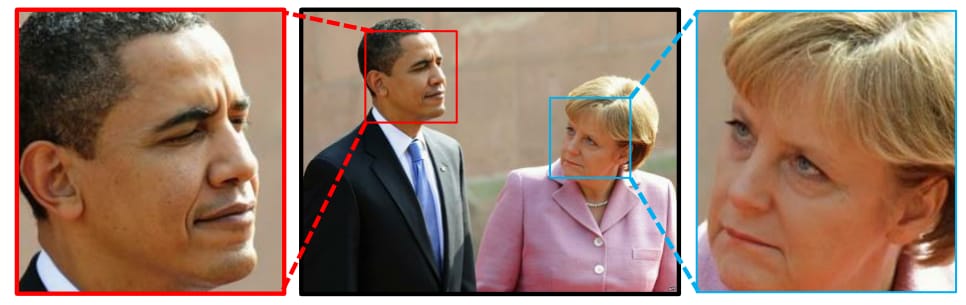}
	\caption{The image is given a caption `\textit{German Chancellor Angela Merkel and U.S. President Barack Obama inspect a military honor guard in Baden-Baden on April 3}.' (source: www.rferl.org). When we examine the face images jointly, we could observe far more rich social facts that are different from that expressed in the text.}
	\label{fig:introduction}
\end{figure}

Interpersonal relation manifests when one establish, reciprocate, or deepen relationships with one another.
The recognition task goes beyond facial expression recognition that analyzes facial motions and facial feature changes of a single subject. It aims for a higher-level interpretation of fine-grained and high-level interpersonal relation traits, such as friendliness, warm, and dominance for faces that co-exist in an image.
Effectively exploiting such relational cues can provide rich social facts. An example is shown in Fig.~\ref{fig:introduction}.
Such a capability promises a wide spectrum of applications. For instance, automatic interpersonal relation inference allows for relation mining from image collection in social networks, personal albums, and films. 
Face-based relational cues can also be combined with other visual cues such as body postures~\cite{chu2015multi} to achieve an even richer modeling and prediction of relations\footnote{Despite we did not study the integration of face and body cues, if body posture and hand gesture information are available, they can be naturally used as additional input channels for our deep models.}.

Profiling unscripted interpersonal relation from face images is non-trivial. Among the most significant challenges are:
\begin{packed_enumerate}
\item Most existing face analysis models only consider a single subject. No existing methods attempt to consider pairwise faces jointly.
\item Relations are governed by a number of high-level facial factors~\cite{Girard:2014,gottman2001facial,hess2000influence}. Thus we need a rich face representation that captures various attributes such as expression, gender, age, and head pose;
\item No single dataset is presently available to encompass all the required facial attribute annotations for learning such a rich representation. In particular, some datasets only contain face expression labels, while other datasets may only be annotated with the gender label. Moreover, these datasets are collected from different environments and exhibit vastly different statistical distributions. Model training on such heterogeneous data remains an open problem.
\end{packed_enumerate}


We address the first problem through formulating a novel deep convolutional network with a Siamese-like architecture~\cite{bromley1994signature}.  
The architecture consists of two convolutional network branches with shared parameters. Each branch is dedicated to one of the faces that co-exist in an image. Outputs of these two branches are fused to allow joint relation reasoning from pairwise faces, where each face serves as the mutual context to the other. 

To address the second challenge, we formulate the convolutional network branches in a multitask framework such that it is capable of learning rich face representation from auxiliary attributes such as head pose, gender, and age, apart from just facial expressions. 
To facilitate the multitask learning, we gather various existing face expression and attribute datasets and additionally label a new large-scale face \textbf{Exp}ression in-the-\textbf{W}ild (ExpW) dataset, which is formed by over $90,000$ web images.

To mitigate the third issue of learning from heterogeneous datasets, we devise a new attribute propagation approach that   
is capable of dealing with missing attribute labels from different datasets, and yet bridging the gap of heterogeneous datasets.
In particular, during the training process, our network dynamically infers missing attribute labels of a sample using Markov Random Field (MRF), conditioned on appearance similarity of that sample with other annotated samples. 
We will show that the attribute propagation approach allows our network to learn effectively from heterogeneous datasets with different annotations and statistical distributions.

The contributions of this study include:
\begin{packed_enumerate}
\item We make the first attempt to investigate face-driven fine-grained interpersonal relation prediction, of which the relation traits are defined based on psychological study~\cite{kiesler1983circle}. We carefully investigate the detectability and quantification of such traits from face image pairs.
\item We formulate a new deep architecture for learning face representation driven by multiple tasks, \eg~pose, expression, and age. Specifically, we introduce a new attribute propagation approach to bridge the gap from heterogeneous sources with potentially missing target attribute labels. We show that this network leads to new state-of-the-art results on widely-used facial expression benchmarks. It also establishes a solid foundation for us to recognize interpersonal relations.
\item We construct a new interpersonal relation dataset labeled with pairwise relation traits supported by psychological studies~\cite{kiesler1983circle,Knutsonfd}. In addition, we also introduce a large-scale facial expression in-the-wild dataset\footnote{Both ExpW and relation datasets are available at \url{http://mmlab.ie.cuhk.edu.hk/projects/socialrelation/index.html}}.
\end{packed_enumerate}

In comparison to our earlier version of this work~\cite{zhang2015learning}, we present a more principle and unified way of addressing the heterogeneous data problem using the MRF-based attribute propagation approach. This is in contrast to the deep bridging layer proposed in our previous work~\cite{zhang2015learning}, which requires external facial alignment step to extract local part appearances for establishing cross-dataset association. In addition, we study more closely on the facial expression recognition problem, which is crucial for accurate interpersonal relation identification. Specifically, we present a new large-scale dataset and conduct extensive experiments against state-of-the-art expression recognition methods. Apart from the methodology, the paper was also substantially improved by providing more technical details and more extensive experimental evaluations.

\section{Related Work}
\label{sec:related_work}

Understanding interpersonal relation can be regarded as a subfield under \textit{social signal processing}~\cite{cristani2013human,pantic2011social,pentland2007social,vinciarelli2009social,vinciarelli2012bridging}, an important multidisciplinary problem that has attracted a surge of interest from computer vision community.
Social signal processing mainly involves facial expression recognition~\cite{zhao2007dynamic,tian2011facial,liu2014learning,ruiz2015emotions,wu2016constrained,fabian2016emotionet,liu2013facial,liu2014deeply,jung2015joint,mollahosseini2016going,fabian2016emotionet,zhao2016peak}. We provide a concise account as follows.

\vspace{0.1cm}
\noindent \textbf{Facial expression recognition.}
A facial expression recognition algorithm usually consists of face representation extraction and classifier construction. Depending on the adopted face representation, existing algorithms can be broadly categorized into two groups: facial action based methods and appearance-based approaches.

Facial action based methods usually exploit the face geometrical information or face action units driven representation for facial expression classification. For example, Tiam~\etal~\cite{tian2001recognizing} use the positions of facial landmarks for facial action recognition and then perform expression analysis. 
Ruiz \etal ~\cite{ruiz2015emotions} combine the tasks of facial action detection and expression recognition to leverage their coherence. 
Liu \etal~\cite{liu2015inspired} construct a deep network to learn a middle representation known as Micro-Action-Pattern (MAP) representation, so as to bridge the semantic gap between low-level features and high-level expression concepts.
Liu \etal~\cite{liu2014deeply} adapt 3D Convolutional Neural Network (CNN) to detect specific facial action parts to obtain discriminative part-based representation.

Appearance-based methods extract features from face patches or the whole face region. A variety of hand-crafted features have been employed, such as LBP~\cite{valstar2012meta,zhao2007dynamic}, HOG~\cite{5771368}, and SIFT~\cite{4813445} features. 
Recently, a number of methods~\cite{jung2015joint,khorrami2015deep,liu2013facial,mollahosseini2016going,Ng:2015:DLE:2818346.2830593,yu2015image,zhao2016peak} attempt to learn facial features directly from raw pixels by deep learning. 
Unlike methods based on hand-crafted features, a deep learning framework allows end-to-end optimization of feature extraction, selection, and expression recognition.
Liu \etal~\cite{liu2013facial} show the effectiveness of Boosted Deep Belief Network (BDBN) for end-to-end feature extraction and selection.
More recent studies~\cite{zhao2016peak} adopt CNN architectures that permit feature extraction and recognition in an end-to-end framework.
For instance, Yu~\etal~\cite{yu2015image} employed an ensemble of multiple deep CNNs. Mollahosseini~\etal~\cite{mollahosseini2016going} used three inception structures~\cite{szegedy2015going} in convolution for facial expression recognition. 
The Peak-Piloted Deep Network (PPDN)~\cite{zhao2016peak} is introduced to implicitly learn the evolution from non-peak to peak expressions.
We introduce readers to a recent survey~\cite{zafeiriou2016facial} focusing on deep learning-based facial behavior analysis.

Our approach is regarded as an appearance-based approach, but differs significantly from the aforementioned studies in that most existing approaches are based on single person, therefore, cannot be directly employed for interpersonal relation inference.
In addition, these studies mostly focus on recognizing prototypical expressions. Interpersonal relation is far more complex involving many factors such as age and gender. Thus we need to consider more attributes jointly in our problem.

\vspace{0.1cm}
\noindent \textbf{Human interaction and group behavior analysis.}
There exists a number of studies that analyze human interaction and group behavior from images and videos~\cite{ding2010learning,ding2011inferring,fathi2012social,ramanathan2013social,ricci2015uncovering,4757440,gallagher2009understanding}.
Many of these studies focus on the coarser level of interpersonal connection other than the one defined by Kiesler in the interpersonal circle~\cite{kiesler1983circle}. For instance, Ding and Yilmaz~\cite{ding2010learning} and  Ricci~\etal~\cite{ricci2015uncovering} only identify the social group (or jointly for estimate head and body orientations) without inferring the relation between individuals.
Fathi \etal~\cite{fathi2012social} only detect three social interaction classes, \ie,~ `dialogue, monologue and discussion'.
Wang \etal~\cite{wang2010seeing} define social relation by several social roles, such as `father-child' and `husband-wife'.
Chakraborty \etal~\cite{chakraborty20133d} classify photos into classes such as `couple, family, group, or crowd'.
Other related problems also include image communicative intents prediction~\cite{6909429} and social role inference~\cite{lan2012social}, usually applied on news and talks shows~\cite{raducanu2012inferring}, or meetings to infer dominance~\cite{hung2007using}.

In comparison to the aforementioned studies~\cite{ding2010learning,fathi2012social}, our work aims to recognize fine-grained and high-level interpersonal relation traits~\cite{kiesler1983circle}, rather than identify social group and roles.
In addition, many of these studies did not use face images directly, but visual concepts~\cite{ding2011inferring} discovered by detectors or people spatial proximity in 2D or 3D spaces~\cite{chen2012discovering}. All these information sources are valuable for learning human interactions but we believe that face still serves a primary role in defining fine-grained and high-level interpersonal relation since face can reveal much richer information such as expression, age, and gender.

Other group behavior studies~\cite{deng2016structure,hoaitalking,ibrahim2016hierarchical,humaninteraction} mainly recognize action-oriented behaviors such as hugging, handshaking or walking, but not face-based interpersonal relations. Often, group spatial configuration and actions are exploited for recognition. Our study differs in that we aim at recognizing abstract relation traits from faces.

\begin{table*}[t]
\caption{A comparison of popular facial expression datasets and the proposed ExpW dataset.}
\label{tab:expression_datasets}
\begin{center}
\begin{tabular}{l|c|c|c|c}
\hline
Datasets&Quantity&Environment&Expression&Data format\\
\hline \hline
JAFFE~\cite{lyons1999automatic}&213 images from 10 subjects&lab&posed&256$\times$256 gray scale image\\
MMI~\cite{1521424}& 238 sequences from 28 subjects &lab& posed&720$\times$576 RGB frames \\
Oulu-CASIA~\cite{zhao2011facial}&480 sequences from 80 subjects&lab&posed& 320$\times$240 RGB frames\\
CK+~\cite{5543262}&593 sequences from 123 subjects&lab&posed& 640$\times$490 or 640$\times$480 gray scale frames\\
FER~\cite{Goodfeli-et-al-2013}&35,587 images&wild&natural& 48$\times$48 gray scale image\\
SFEW~\cite{Dhall:2015:VIB:2818346.2829994}&1,635 images&wild&natural&720$\times$576 RGB images\\
ExpW&91,793 images&wild&natural& Original web images\\
\hline
\end{tabular}
\end{center}
\vspace{-0.2cm}
\end{table*}

\vspace{0.1cm}
\noindent \textbf{Deep learning.}
Deep learning has achieved remarkable success in many tasks of face analysis, \eg~face detection~\cite{yang2015facial,yang2016wider,li2015convolutional,yang2015convolutional,opitz2016grid}, face parsing \cite{pluo2,liu2015multi}, face landmark detection~\cite{zhang,zhu2015face,trigeorgis2016mnemonic}, face attribute recognition \cite{ziwei,wang2016walk,huang2016learning}, face recognition \cite{schroff2015facenet,parkhi2015deep,sun2016sparsifying}, and face clustering~\cite{zhang2016joint}.
However, deep learning has not yet been adopted for face-driven interpersonal relation mining that requires joint reasoning from multiple persons. In this work, we propose a deep model to capture complex facial attributes from heterogeneous datasets, and joint learning from face pairs. Although there are several algorithms~\cite{bi2014multilabel,yu2014large,yang2016improving} that perform training on heterogeneous datasets, most of these studies assume fixed image features and exploit the label correlation for missing label propagation. Lee~\cite{lee2013pseudo} proposes a deep learning algorithm that employs pseudo label to utilize the unlabeled data. But the pseudo label is simply generated by a pre-trained network using labeled data, thus the potential correlation between the labeled and unlabeled data is ignored.
Our network also differs from the multitask network in~\cite{zhang}, which assumes complete labels from all attributes and homogeneous data sources.

\section{Face Expression and Interpersonal Relation Datasets}
\label{sec:datesets}

Before we describe our approach, we introduce two new datasets collected in this study.
\subsection{Face Expression Dataset}
\label{subsec:expression_dataset}

Research in face perception and emotion typically requires very large annotated datasets of images of facial expressions. 
There are a number of facial expression datasets, \eg, CK+~\cite{5543262}, JAFFE~\cite{lyons1999automatic}, Oulu-CASIA~\cite{zhao2011facial}, MMI~\cite{1521424}, FER \cite{Goodfeli-et-al-2013}, SFEW \cite{Dhall:2015:VIB:2818346.2829994}. A summary is provided in Table~\ref{tab:expression_datasets}. These datasets are either collected in controlled environments, or the quantity is insufficient to train a robust deep network. An automatic method for expression dataset construction is proposed in~\cite{fabian2016emotionet}. This method is useful to collect large-scale dataset. Nonetheless, it relies on accurate facial landmark detection and thus may limit face variations in the collected data.  

\begin{figure}[b]
	\centering
	\includegraphics[width=1\linewidth]{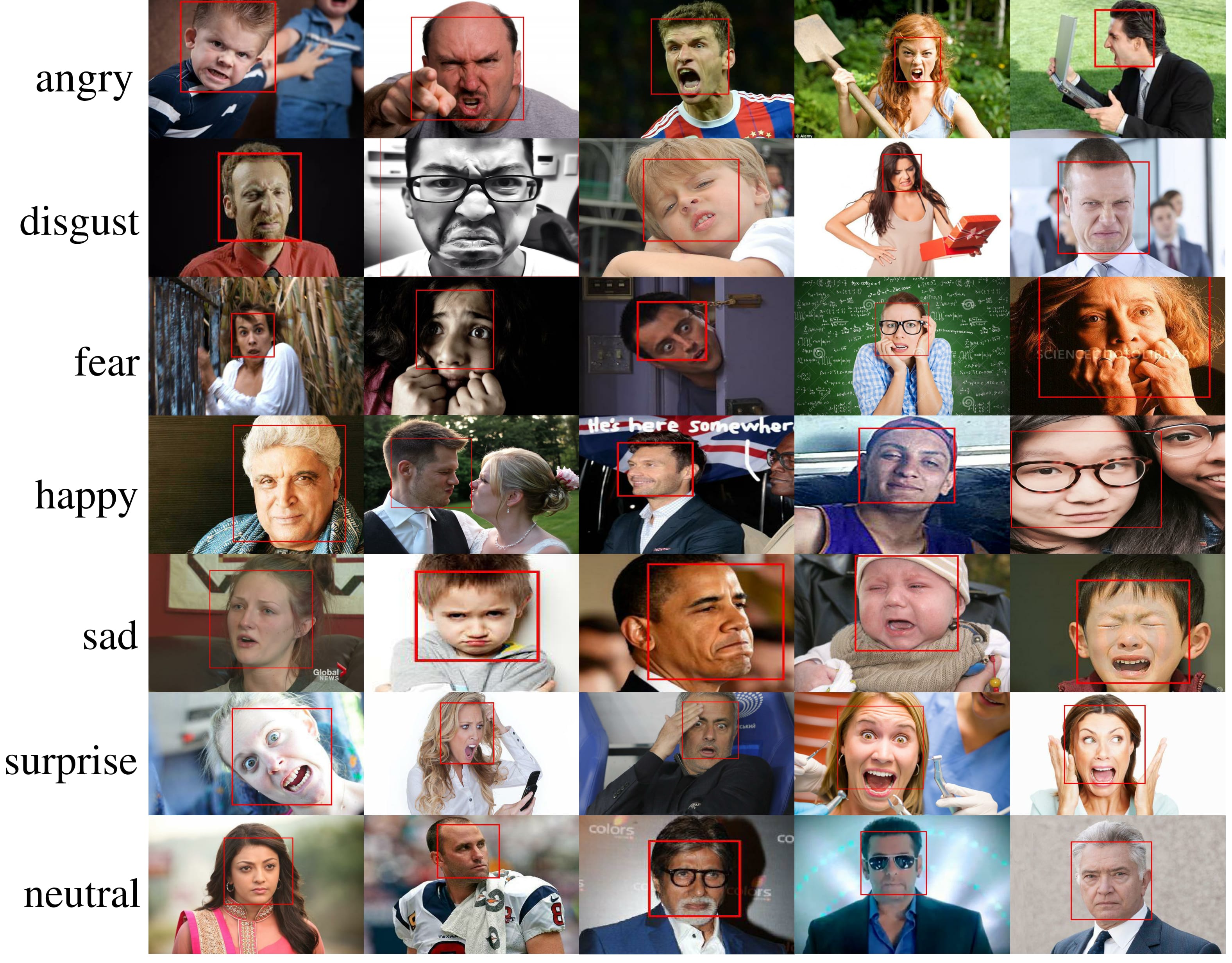}
	\caption{Example images of the proposed ExpW dataset.}
	\label{fig:expw_example}
\end{figure}

To this end, we built a new database named as Expression in-the-Wild (ExpW) dataset that contains 91,793 faces manually labeled with expressions.
The quantity of images in ExpW is larger and the face variations are more diverse than many existing databases, as summarized in Table~\ref{tab:expression_datasets}. Figure~\ref{fig:expw_example} shows some example images of ExpW.

We collected ExpW dataset in the following way.
Firstly, we prepared a list of emotion-related keywords such as ``excited'', ``afraid'' and ``panic''. Then we appended different nouns related to a variety of occupations, such as ``student'', ``teacher'', and ``lawyer'', to these words and used them as queries for Google image search. 
Subsequently, we collected images returned from the search engine and run a face detector~\cite{yang2014aggregate} to obtain face regions from these images. Similar to other existing expression datasets~\cite{Dhall:2015:VIB:2818346.2829994,Goodfeli-et-al-2013}, each of the face images was manually annotated as one of the seven basic expression categories: ``angry'', ``disgust'', ``fear'', ``happy'', ``sad'', ``surprise'', or ``neutral''. Non-face images were removed in the annotation process.


\subsection{Interpersonal Relation Dataset}
\label{subsec:social_relation_dataset}

\begin{figure*}[t]
	\centering
	\includegraphics[width=1\linewidth]{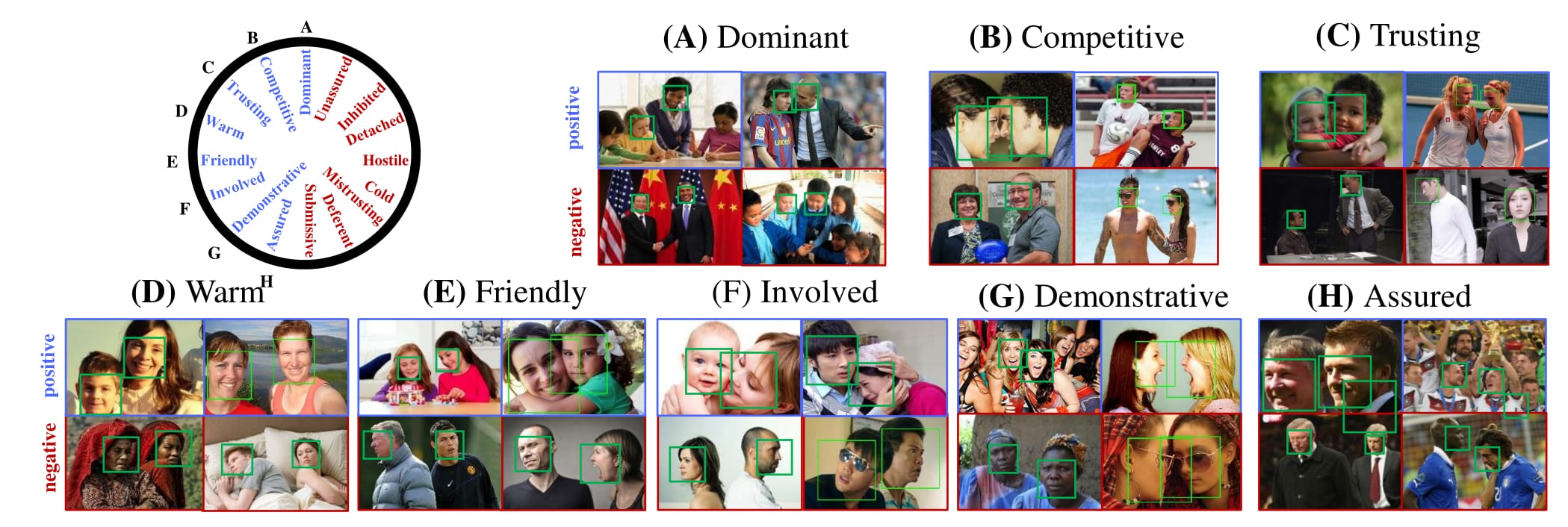}
	\vskip -0.25cm
	\caption{The 1982 Interpersonal Circle (upper left) is proposed by Donald J. Kiesler, and commonly used in psychological studies~\cite{kiesler1983circle}. The 16 segments in the circle can be grouped into 8 relation traits. The traits are non-exclusive therefore can co-occur in an image. In this study, we investigate the detectability and quantification of these traits from computer vision point of view. (A)-(H) illustrate positive and negative examples of the eight relation traits.}
	\label{fig:relation_definition}
\end{figure*}

To investigate the detectability of relation traits from a pair of face images, we built a new dataset containing $8,016$ images chosen from web and movies. Each image was labeled with faces' bounding boxes and their pairwise relations. This is the first face dataset annotated with interpersonal relation traits. 
It is challenging because of large face variations including poses, occlusions, and illuminations. In addition, the images exhibit rich relation traits from various sources including news photos of politicians, photos in social media, and video frames in movies, as shown in Fig.~\ref{fig:relation_definition}.

\begin{table*}[t]
	\newcommand{\tabincell}[2]{\begin{tabular}{@{}#1@{}}#2\end{tabular}}
	\caption{Descriptions of interpersonal relation traits based on the 1982 interpersonal circle~\cite{kiesler1983circle}.}
	\label{tab:social_relation_def}
	\setlength{\tabcolsep}{.35667em}
	\begin{center}
		\begin{tabular}{l|c|c}
			\hline
			Relation trait & Descriptions & Example pair\\
			\hline
			\hline
			Dominant& \tabincell{c}{one leads, directs, or controls the other /\\ dominates the conversation / gives advices to the other} & teacher \& student\\
			\hline
			Competitive& hard and unsmiling / contest for advancement in power, fame, or wealth & people in a debate\\
			\hline
			Trusting& \tabincell{c}{sincerely look at each other / no frowning or showing doubtful expression / \\not-on-guard about harm from each other} &partners\\
			\hline
			Warm& speak in a gentle way / look relaxed / readily to show tender feelings & mother \& baby\\
			\hline
			Friendly& work or act together / express sunny face / act in a polite way / be helpful &host \& guest\\
			\hline
			Involved& engaged in physical interaction / involved with each other / not being alone or separated & lovers\\
			\hline
			Demonstrative& \tabincell{c}{talk freely being unreserved in speech /  \\readily to express the thoughts instead of keep silent / act emotionally}&friends in a party\\
			\hline
			Assured& express to each other a feeling of bright and positive self-concept, instead of depressed or helpless& teammates\\
			\hline
		\end{tabular}
	\end{center}
	\vspace{-0.2cm}
\end{table*}

\begin{table*}[t]
\caption{Example adjectives for relation traits defined by Donald J. Kiesler~\cite{kiesler1983circle}.}
\label{tab:adjectives}
\begin{center}
\begin{tabular}{l|c|c}
\hline
Relation trait&Positive&Negative\\
\hline \hline
dominant&controlling/leading/influencing/commanding/dictatorial&equal/matched/\\

competitive&critical/driven/enterprising&content/approving/flattering/respectful\\

trusting&unguarded/generous/innocent&mistrusting/suspicious/cunning/vigilant\\

warm&gentle/pardoning/soft/absolving&cold/strict/icy/harsh/cruel\\

friendly&cooperative/helpful/devoted&hostile/harmful/impolite/rude\\

involved&outgoing/attached/active/sociable&detached/distant/aloof\\

demonstrative&talkative/casual/suggestive&mute/controlled/silent/unresponsive\\

assured&confident/cheerful/self-reliant/cocky&dependent/unassured/helpless/depressed\\
\hline
\end{tabular}
\end{center}
\end{table*}

Before we collected for annotations, we first defined the interpersonal relation traits based on the interpersonal circle proposed by Kiesler~\cite{kiesler1983circle} that commonly used in psychological studies, where human relations are divided into 16 segments as shown in Fig.~\ref{fig:relation_definition}. Each segment has its opposite side in the circle, such as ``friendly and hostile''. Therefore, the 16 segments can be considered as eight binary relation traits, whose descriptions~\cite{kiesler1983circle} and examples are given in Table~\ref{tab:social_relation_def}. We also provide positive and negative visual samples for each relation in Fig.~\ref{fig:relation_definition}, showing that they are visually perceptible. For instance, ``friendly'' and ``competitive'' are easily separable because of the conflicting meanings. It is worth pointing out that some relations are close semantically, such as ``friendly'' and ``trusting''. To accommodate such cases, we do not forcefully suppress any one of these relations during prediction but allowing a pair of faces to have more than one relation.

\begin{table*}[t]
	\newcommand{\tabincell}[2]{\begin{tabular}{@{}#1@{}}#2\end{tabular}}
	\caption{A summary of attributes annotated in AFLW~\cite{6130513}, CelebA~\cite{ziwei} and the proposed ExpW datasets, each of which contains 24,386, 202,599, and 91,793 face images, respectively.}
	\vskip -0.25cm
	\footnotesize
	\label{tab:attribute_datasets}
	\begin{center}
\setlength{\tabcolsep}{.4667em}
		\begin{tabular}{l|c|c|c|c|c|c|c|c|c|c|c|c|c|c|c|c|c|c|c|c|c|c|c|c}
			\hline
			 \multirow{2}[10]{*}{Attributes}&Gender&\multicolumn{5}{c|}{Pose}&\multicolumn{10}{c|}{Expression}&\multicolumn{8}{c}{Age}\\\cline{2-25}
			&\rotatebox{90}{gender}&\rotatebox{90}{left profile}&\rotatebox{90}{left}&\rotatebox{90}{frontal}&\rotatebox{90}{right}&
			\rotatebox{90}{right profile}&
			\rotatebox{90}{angry}&\rotatebox{90}{disgust}&\rotatebox{90}{fear}&
			 \rotatebox{90}{happy}&\rotatebox{90}{sad}&\rotatebox{90}{surprise}&\rotatebox{90}{neutral}&\rotatebox{90}{smiling}&
			\rotatebox{90}{\parbox{1.1cm}{mouth \\ opened}}&
			\rotatebox{90}{\parbox{1.1cm}{narrow \\ eyes}}&
			\rotatebox{90}{young}&\rotatebox{90}{goatee}&
			\rotatebox{90}{no beard}&\rotatebox{90}{sideburns}&
			\rotatebox{90}{\parbox{1.5cm}{5 o'clock \\ shadow}}&
			\rotatebox{90}{gray hair}&
			\rotatebox{90}{bald}&
			\rotatebox{90}{mustache}\\
			\hline\hline
			AFLW~\cite{6130513}&$\surd$&$\surd$&$\surd$&$\surd$&$\surd$&$\surd$&&&&&&&&&&&&&&&&&\\
			
			 CelebA~\cite{ziwei}&$\surd$&&&&&&&&&&&&&$\surd$&$\surd$&$\surd$&$\surd$&$\surd$&$\surd$&$\surd$&$\surd$&$\surd$&$\surd$&$\surd$\\
			
			 \tabincell{l}{ExpW}&&&&&&&$\surd$&$\surd$&$\surd$&$\surd$&$\surd$&$\surd$&$\surd$&&&&&&&&&&\\
			\hline
		\end{tabular}
	\end{center}
\end{table*}

Annotating relations is non-trivial and subjective by nature. We requested five performing arts students to label each relation for each face pair independently. A label was accepted if more than three annotations were consistent. The inconsistent samples were presented again to the five annotators to seek for consensus.
To facilitate the annotation task, we also provided multiple cues to the annotators. First, to help them understand the definition of the relation traits, we listed ten related adjectives (see Table~\ref{tab:adjectives} for examples) defined by ~\cite{kiesler1983circle} for the positive and negative samples on each relation trait, respectively. Multiple example images were also provided. Second, for image frames selected from movies, the annotators were asked to get familiar with the plot. The subtitles were presented during the labeling process. Third, we defined some measurable rules for the annotation of all relation traits. For example, if two people open their mouths, the relation trait of ``demonstrative'' is considered as positive; If a teacher is teaching his student, the ``dominant'' trait is considered as positive; A trait is defined as negative if the annotator cannot find any evidence to support its positive existence. 
The average Fleiss' kappa of the eight relation traits annotation is 0.62, indicating substantial inter-rater agreement.

\section{Facial Expression and Attributes Recognition}
\label{sec:exp_representation}

The recognition of facial expression and other relevant attributes such as gender and age play a critical role in our relation prediction framework.
In this study, we train a deep convolutional network end-to-end to map raw imagery pixels to a representation space and then perform expression and attribute prediction simultaneously. The joint learning of facial expression and attributes allows us to capture rich facial representation more effectively thus preparing a strong starting point for interpersonal relation learning. 

\subsection{Problem Formulation and Overview}

A natural way to learn a deep representation that captures multiple attributes is by training a multitask network that jointly predicts these attributes given a face image~\cite{zhang}. This can be implemented directly by introducing multiple supervisory tasks during the network training.
In our problem training a multitask network, unfortunately, is non-trivial:
\begin{enumerate}
\item \textit{Missing attribute labels} - As discussed in Section~\ref{sec:intro}, face datasets that can cover all different kinds of attributes can hardly be found.
The ExpW dataset collected by us, and the few popular face datasets such as AFLW~\cite{6130513} and CelebA~\cite{ziwei} contain subsets of attributes useful for our problem, but these subsets rarely overlap, as shown in Table~\ref{tab:attribute_datasets}. For instance, AFLW only contains gender and poses, while the ExpW dataset only has expressions. 
The many missing labels prevent us from `fully' exploit an image since it is labeled with an attribute subset rather than a complete attribute set. 
The problem may also lead to sparsity in the supervisory signal and thus increase the convergence difficulty during training.

\item \textit{Heterogeneous distribution} - These datasets were collected from different sources, therefore, exhibit vastly disparate statistical distributions. Specifically, the AFLW dataset contains face images gathered from Flickr that typically hosts high-quality photographs. Whereas the image quality in CelebA and ExpW is much lower and more diverse. Since these datasets are labeled with different sets of attributes, a direct joint training would bias each attribute to be trained by the corresponding labeled data alone, instead of benefiting from the existence of unlabeled images.

\end{enumerate}

\begin{figure}[t]
	\centering
	\includegraphics[width=1\linewidth]{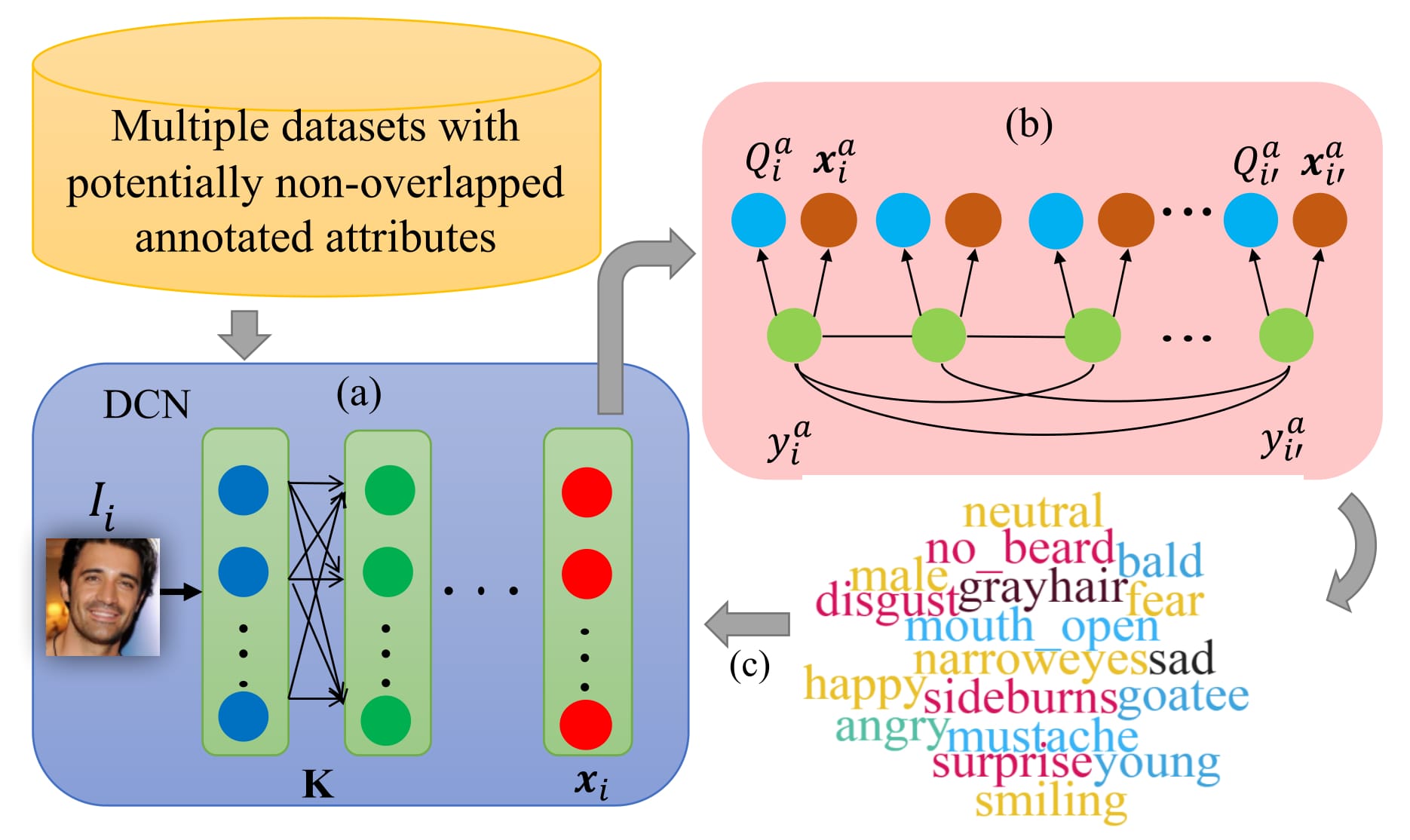}
	\caption{An illustration of the second stage training, in which we perform alternating optimization of representation learning and attribute propagation. (a) We extract face representation $\mathbf{x}_i$ from the initialized DCN. (b) Given the face representation and attribute correlation, we perform attribute propagation in a Markov Random Field (MRF) to infer the missing attribute labels. (c) We refine the DCN by  using the ground truth labels and pseudo labels generated from MRF.}
	\label{fig:attribute_network}
\end{figure}

We propose a novel learning framework to mitigate the aforementioned problems. In general, given the training faces from multiple heterogeneous sources, we aim to train a deep convolutional network (DCN) that can predict the \textit{union set} of attributes of these datasets (\ie~all attributes in Table~\ref{tab:attribute_datasets}).
The training process is divided into two stages, as summarized in Algorithm~\ref{alg:Framework}. Further details of each stage are provided in Sec.~\ref{subsec:network_initialization} and Sec.~\ref{subsec:attribute_propagation}.
\begin{enumerate}
\item \textit{Network initialization} -  
Firstly, we initialize the parameters of our deep convolutional network by training it to minimize the classification error on the attributes despite the missing attribute labels in some samples.   

\item \textit{Alternating attribute propagation and face representation learning} - 
We fine-tune the network from the first stage via an alternating optimization process for obtaining a better face representation. The process is depicted in Fig.~\ref{fig:attribute_network}.
In each iteration of the optimization, we extract the deep representation from each face, and compute the prior of attribute co-occurrence, based on which we perform attribute propagation to infer the missing attribute annotations as pseudo attribute labels in a MRF. We subsequently refine the network supervised by the ground truth attribute labels and newly generated pseudo attribute labels. 

\end{enumerate}

\begin{algorithm}[t]
\caption{Overview of the proposed framework.}
\label{alg:Framework}
\begin{algorithmic}[1]
\REQUIRE ~~\\
Multiple face image datasets with potentially non-overlapped attribute annotations.
 \ENSURE ~~\\
Face representation that captures the union of the attributes from input datasets.

\hspace{-0.5cm} \textbf{Stage 1 Training:} \\
\STATE Initialize the network filters $\K$ by maximizing Eqn.~(\ref{eq:init_cnn}).

\hspace{-0.5cm} \textbf{Stage 2 Training:} \\
\FOR {$m$ = 1 to $M$}
\STATE Perform attribute propagation to fill up the missing labels by maximizing Eqn.~(\ref{eq:MRF1}).
\STATE Refine the network filters $\K$ supervised by the ground truth and pseudo labels by minimizing the attribute classification error. 
\ENDFOR
\end{algorithmic}
\end{algorithm}

The second stage of Algorithm~\ref{alg:Framework} helps to provide pseudo attribute labels that are missing initially for network fine-tuning. There are two advantages of this method: 1) The attribute propagation process does not require any prior knowledge of the problem at hand and thus can be applied given other datasets with an arbitrary number of missing labels. 2) Filling up the missing labels with pseudo labels naturally establish shared tasks among the datasets and gradually bridge the gap between datasets of different distributions.
We show in the experiments (Sec.~\ref{sec:experiments}) that pseudo labels obtained in the attribute propagation step are crucial for good performance in the task of relation prediction.


\subsection{First Stage: Network Initialization}
\label{subsec:network_initialization}

The first stage of our training process is network initialization.
Specifically, we first train the DCN (Fig.~\ref{fig:attribute_network}(a)) using a combined dataset, which includes AFLW, CelebA, and ExpW.
Note that we do not perform attribute propagation at this stage but allow missing labels in samples.

Formally, let the network parameters be $\K$, an input face image $\I_i$ is transformed to a higher level of representation represented as $\x_i=\Phi(\I_i|\K)$, where $\Phi(\I_i|\K)$ denotes a nonlinear mapping parameterized by $\K$. 
We employ the Batch-Normalized Inception architecture presented in~\cite{ioffe2015batch}, where the network input is 224$\times$224 RGB image, and the generated face representation $\x_i\in \mathbb{R}^{1024\times1}$.

We assume the attributes are binary and thus we compute the probability for an attribute $a$ by logistic regression. More precisely, given the attribute label $y^a\in\{0,1\}$, we have $p(y^a=1|\I;\K,\w^{a})=\frac{1}{1+\exp(-\w^{a}\Phi(\I_i|\K))}$, where $\w^{a}$ are parameters of the logistic classifier. The network filter $\K$ and classifier parameter $\w^a$ can be obtained by maximizing the posterior probability:
\begin{equation}\label{eq:init_cnn}
\K^*,\W^{A^*}=\argmax_{\W^A,\K} \sum_{i=1}^{N}\sum_{a=1}^{|A|}\log p(y^a_i|\I_i;\w^a,\K),
\end{equation}
where $N$ is the number of training samples, $y^a_i$ is the ground truth label, $A$ denotes the set of attributes, and $\W^A=\{\w^a\}_{a \in A}$. 
As a result, we can formulate a loss function with cross entropy for each attribute. The training process is conducted via back-propagation (BP) using stochastic gradient descent (SGD)~\cite{krizhevsky2012imagenet}. 

Note that there are missing labels in the training set, which is combined from arbitrary datasets. To mitigate this issue, we mask the error of the missing attribute $a$ of a training sample, and only back-propagate errors if the ground truth label of an attribute exists.
%
%
Despite the missing labels, this simple approach provides a good initialization point for the second stage of the training.

\begin{algorithm}[b]
\caption{Alternating attribute propagation and face representation learning.}
\label{alg:label_propogation}
\begin{algorithmic}[1]
\REQUIRE ~~\\
Face representation $\x$, and datasets with partially labeled attributes.
 \ENSURE ~~\\
Pseudo label $\Y^a$ on an attribute $a$ for unlabeled data.

\STATE Compute the attribute co-occurrence prior $\Q^a$ and extract face representation $\X$.
\STATE For labeled data, use the original annotations; For unlabeled data, initialize the label by K-NN classification using the labeled data. Then we have the initial pseudo label $\Y^a$.
\STATE Initialize the model parameter $\Omega^a$ in Eqn.~(\ref{eq:unary_potential}).
\STATE Compute the affinity matrix $V$ in Eqn.~(\ref{eq:pairwise_potential}).
\STATE Let iteration $t=0$.
\WHILE {not converged}
\STATE $t=t+1$.
\STATE Infer a new $\Y^a_t$ given the face representation $\X$ and current model parameter $\Omega^a_t$ (Eqn.~(\ref{eqn:1})-(\ref{eqn:3})). Set $\Y^a=\Y^a_t$.
\STATE Update $\Omega^a_t$ to maximize the log-likelihood of $p(\X,\Y^a,\Q^a)$ by EM algorithm (Eqn.~(\ref{eqn:4})-(\ref{eqn:5})).
\ENDWHILE
\end{algorithmic}
\end{algorithm}

\subsection{Second Stage: Alternating Attribute Propagation and Face Representation Learning}
\label{subsec:attribute_propagation}

\noindent\textbf{Formulation}. Following Algorithm~\ref{alg:Framework}, with the initialized network parameter $\K$ and attribute classifier parameters $\W^A$, we subsequently perform attribute propagation to infer the missing attributes.

%

Attribute propagation is achieved based on two criteria: 1) Similarity of appearances between two faces, and 2) the correlation between attributes.
The first criterion implies that the attributes of two faces are likely the same if their facial appearances are close to each other. 
The second criterion reflects the fact that some attributes, such as `happy' and `smiling', often co-occur. 

With the above intuition, we formulate the attribute propagation problem in a MRF framework.
In particular, as depicted in Fig.~\ref{fig:attribute_network}(b), each node in the MRF is an attribute label $y^a_i$ for an image sample $\I_i$. Each edge describes the relation between the labels. For each node, we associate it with the observed variables $\x_i$ representing the face representation obtained from the DCN, and $Q^{a,a'}_i$, which serves as a co-occurrence prior that indicates the tendency of an attribute $a$ is present on a face $i$, given another attribute $a'$ as condition. 

We first provide the definition of the co-occurrence prior $Q^{a,a'}_i$.
Given an attribute $a$ and another attribute $a'\in A\backslash a$, we define $Q^{a,a'}_i$ as
\begin{equation}
Q^{a,a'}_i=
	\begin{cases}
		\frac{cov(\w^a,\w^{a'})}{\sigma_{\w^a}\sigma_{\w^{a'}}}  &\text{if~} y^{a'}_i=1~(\text{positive})\\
		-\frac{cov(\w^a,\w^{a'})}{\sigma_{\w^a}\sigma_{\w^{a'}}}  &\text{if~} y^{a'}_i=0~(\text{negative})\\
		0 & \text{if $y^{a'}_i$ is unlabeled}.
	\end{cases}
\end{equation}
More precisely, $Q^{a,a'}_i$ is assigned with the Pearson product-moment correlation coefficient~\cite{pearson1895note}, of which the sign is governed by the ground truth label of attribute $a'$, \ie~$y^{a'}_i$. The $cov(\cdot)$ is the covariance, and $\sigma$ is the standard deviation, while $\w^a$ and $\w^{a'}$ represent the parameters of the logistic classifier for the respective attribute.
Intuitively, if attributes $a$ and $a'$ tend to co-occur, their $\w^a$ and $\w^{a'}$ are positively correlated. 
For instance, we have $a=$``happy'', $a'=$``smiling'', and the Pearson correlation $\frac{cov(\w^a,\w^{a'})}{\sigma_{\w^a}\sigma_{\w^{a'}}}=0.3$. 
For a face $i$, if the attribute ``smiling'' is annotated as positive (\ie~$y^{a'}_i=1$), then we have $Q^{a,a'}_i=0.3$, suggesting that the ``happy'' attribute is present on the face given the ``smiling'' attribute.
On the contrary, if the attribute ``smiling'' is absent (\ie~$y^{a'}_i=0$), then $Q^{a,a'}_i=-0.3$, suggesting that the ``happy'' attribute is likely to absent too.
We treat unannotated $y^{a'}_i$ as a special case by forcing $Q^{a,a'}_i=0$.

Let the face representation $\X=\{\x_i\}$ and attribute co-occurrence prior $\Q^a=\{Q^{a}_i\}$, we maximize the following joint probability to obtain the attribute labels $\Y^a=\{y_i^a\}$:
\begin{equation}\label{eq:MRF1}
\begin{split}
p(\X,\Y^a,\Q^a) &= p(\X,\Q^a|\Y^a)p(\Y^a)\\
&=\frac{1}{Z}\prod_{i}\Phi(\x_i,Q^a_i|y^a_i)\prod_{i}\prod_{i'\in \mathcal{N}_i}\Psi(y^a_i,y^{a}_{i'})
\end{split}
\end{equation}
where $\Phi(\cdot)$, $\Psi(\cdot)$ is the unary and pairwise term, respectively. The $Z$ is the partition function, and $\mathcal{N}_i$ denotes a set of face images, which are the neighbors of $y^a_i$. 
%
%

We explain the unary and pairwise terms of Eqn.~(\ref{eq:MRF1}) as follows:

\vspace{0.1cm}
\noindent \textit{Unary term} - 
We employ the Gaussian distribution to model the feature $\x_i$ in the unary term $\Phi(\cdot)$. 
And we use the attribute co-occurrence prior as the prior probability. Specifically, 
\begin{equation}\label{eq:unary_potential}
\Phi(\x_i,Q^a_i|y^a_i=\ell)\thicksim\mathcal{N}(\x_i|\mu^a_{\ell},\Sigma^a_{\ell})\cdot\prod_{a'\in A\backslash a}\mathcal{S}_{y^a_i}(Q^{a,a'}),
\end{equation}
where $\ell \in \{0,1\}$, $\mu_{\ell}$ and $\Sigma_{\ell}$ denote the mean vector and covariance matrix of samples when $y^a_i=\ell$. Both $\mu_{\ell}$ and $\Sigma_{\ell}$ are obtained and updated during the inference process. For simplicity, we denote the model parameter $\Omega^a=\{\mu^a_{\ell}, \Sigma^a_{\ell}\}$ in the following text.  
For $\mathcal{S}_{y^a_i}(Q^{a,a'}_i)$, recall that given the attribute $a'$, $Q^{a,a'}$ denotes the prior that attribute $a$ appears. Here we define $\mathcal{S}_{y^a_i}$ as:

\begin{equation}
\mathcal{S}_{y^a_i}(Q^{a,a'}_i)=
	\begin{cases}
\text{sigmoid}(Q^{a,a'}_i)  &y^a_i=1, \\
1-\text{sigmoid}(Q^{a,a'}_i)  &y^a_i=0.\\
	\end{cases}
\end{equation}
Here ``sigmoid'' denotes a sigmoid transformation that maps the attribute co-occurrence prior from the range of [-1,1] to [0,1]. Hence, $\mathcal{S}_{y^a_i}(Q^{a,a'}_i)$ describes the prior that the attribute appears ($y^a_i=1$) or not ($y^a_i=0$).

\vspace{0.1cm}
\noindent \textit{Pairwise term} - 
The pairwise term $\Psi_p(\cdot)$ in Eqn.(\ref{eq:MRF1}) is defined as
\begin{equation}
\label{eq:pairwise_potential}
\begin{split}
\Psi(y^a_i,y^{a}_{i'})=\exp\{v_{ii'}\cdot \text{sign}(y^a_i,y^{a}_{i'})\},
\end{split}
\end{equation}
where $\text{sign}(\cdot)$ denotes a sign function:
\begin{equation}
 \text{sign}(y^a_i,y^{a}_{i'})=
	\begin{cases}
1  &y^a_i=y^{a}_{i'}\\
		-1&\text{otherwise}.
	\end{cases}
\end{equation}
The variable $v_{ii'}$ encodes the affinity between arbitrary pair of face image features $\x_i$ and $\x_{i'}$. 
We obtain $v_{ii'}$ via the spectral clustering approach presented in~\cite{zelnik2004self}. 

Firstly, we compute an affinity matrix $V$ with entries $v_{ii'} = \exp(-d^2(\x_{i},\x_{i'})/\sigma_i\sigma_{i'})$ if $\x_{i'}$ is within the $h$-nearest neighbors of $\x_i$, otherwise we set $v_{ii'}=0$. We set $h=10$ in this study. 
The term $d(\x_i,\x_{i'})$ is the $\ell_2$-distance between $\x_i$ and $\x_{i'}$, and $\sigma_i$ is the local scaling factor with $\sigma_i = d(\x_i,\x_h)$, where $\x_h$ is the $h$-th nearest neighbor of $\x_i$. Then the normalized affinity matrix is obtained by $V=D^{-\frac{1}{2}}VD^{-\frac{1}{2}}$, where $D$ is a diagonal matrix with $D_{ii'} = \sum_{i'=1}^{n}v_{ii'}$. 
%
%
Intuitively, Eqn.~(\ref{eq:pairwise_potential}) penalizes face images with high affinity to be assigned with different attribute labels. 


\vspace{0.2cm}
\noindent\textbf{Optimization}. Given the face representation $\X$ and attribute co-occurrence prior $\Q^a$, we infer the missing attribute labels $\Y^a$ by maximizing the joint probability of Eqn.~(\ref{eq:MRF1}).


Firstly, for the unlabeled data, we initialize the attribute $a$ by K-NN classification in the space of $\x$ using the labeled data. We keep the original attribute annotations for labeled data. Then we obtain $\mu^a_{\ell}$ and $\Sigma^a_{\ell}$ from the Gaussian of samples with $y^a=\ell$. 

%
%
%

After the initialization of $\Omega^a$, we infer $\Y^a$ and update the model parameter $\Omega^a$ by repeating the following two steps in each optimization iteration $t$:
\begin{enumerate}
	 \item Infer a new $\Y^a_t$ given the face representation $\X$ and model parameter $\Omega^a_t$. Set $\Y^a = \Y^a_t$.
  \item Given $\Y^a$, update $\Omega^a_t$ to maximize the log-likelihood of $p(\X,\Y^a,\Q^a)$ by Expectation-Maximization (EM) algorithm.
\end{enumerate}

For the first step, we aim to obtain a new $\Y^a_t$ given $\X$ and model parameter $\Omega^a_t$. A natural way is to infer from the posterior:
 \begin{equation}
 \label{eqn:1}
 p(\Y^a|\X,\Q^a,\Omega^a_t)=\frac{p(\X,\Q^a|\Y^a,\Omega^a_t)p(\Y^a)}{p(\X,\Q^a|\Omega^a_t)}.
 \end{equation}
 However the computation of the term $p(\Y^a)$ involves the interaction of each $y^a_i$ and its neighborhood (\ie~the $h$-nearest neighbors in the space of $\x$). Thus, it is intractable. Here we employ the mean field-like approximation~\cite{celeux2003procedures} for $p(\Y^a)$ computation, in which we assume each $y^a_i$ is independent, and we set the value of its neighborhood $\mathcal{N}_i$ constant when we compute $p(y_i)$. In this case, we have
 \begin{equation}
  \label{eqn:2.0}
 p(\Y^a) = \prod_{i}p(y^a_i|\mathbb{Y}^a_{\mathcal{N}_i}),
 \end{equation}
 where we denote the value of $y_i$'s neighborhood as $\mathbb{Y}^a_{\mathcal{N}_i}\in\mathbb{R}^{|\mathcal{N}_i|\times 1}$. For example, we can reuse the value in the previous iteration $t-1$ (\ie~$\mathbb{Y}^a_{\mathcal{N}_i}=\Y^a_{(t-1)\mathcal{N}_i}$). Because $y^a_i\in\{0,1\}$, we have 
 \begin{equation}
  \label{eqn:2}
  \begin{split}
 p(y^a_i|\mathbb{Y}^a_{\mathcal{N}_i})&=\frac{p(y_i,\mathbb{Y}^a_{\mathcal{N}_i})}{\sum_{y_i\in\{0,1\}}p(y_i,\mathbb{Y}^a_{\mathcal{N}_i})}\\
 & = \frac{\frac{1}{Z}\prod_{j\in\mathcal{N}_i}\Psi(y^a_i,y^a_j)}{\sum_{y_i\in\{0,1\}}\frac{1}{Z}\prod_{j\in\mathcal{N}_i}\Psi(y^a_i,y^a_j)}.
 \end{split}
 \end{equation}
 Since $\mathbb{Y}^a_{\mathcal{N}_i}$ is fixed, the partition function $Z$ is constant when we compute $p(y_i,\mathbb{Y}_{\mathcal{N}_i})$. Thus $Z$ can be eliminated in Eqn.~(\ref{eqn:2}). Combining Eqn.~(\ref{eq:unary_potential}), Eqn.~(\ref{eqn:1}), Eqn.~(\ref{eqn:2}), we have
 \begin{equation}
 \begin{split}
 &p(\Y^a|\X,\Q^a,\Omega^a_t)=\prod_{i}p(y^a_i|\mathbb{Y}^a_{\mathcal{N}_i},\x_i,Q^a_i,\Omega^a_t)\\
&=\prod_{i}\frac{\Phi(\x_i,Q^a_i|y^a_i,\Omega^a_t)\prod\limits_{j\in\mathcal{N}_i}\Psi(y^a_i,y^a_j)}{\sum\limits_{y^a_i\in\{0,1\}}\Phi(\x_i,Q^a_i|y^a_i,\Omega^a_t)\prod\limits_{j\in\mathcal{N}_i}\Psi(y^a_i,y^a_j)}.
\label{eqn:3}
\end{split}
 \end{equation}
 Intuitively, the posterior $p(y^a_i=\ell|\mathbb{Y}^a_{\mathcal{N}_i},\x_i, \Omega^a_t)$ is proportional to the likelihood of setting $y^a_i=\ell$, with the neighborhood's value fixed. Then this posterior can be computed directly for each face $i$. To this end, we have 
\begin{equation}
\Y^a_t=\{y_i^a\}_{i=1}^{N}
\end{equation}
 where for the unlabeled samples, $y^a_i$ is simulated based on the posterior $p(y^a_i=\ell|\mathbb{Y}^a_{\mathcal{N}_i},\x_i, \Omega^a_t)$ (\ie~the probability of setting $y^a_i=\ell$ is proportional to $p(y^a_i=\ell|\mathbb{Y}^a_{\mathcal{N}_i},\x_i, \Omega^a_t)$). For the annotated samples, we use the annotation directly.

For the second step, we aim to maximize the log-likelihood of $p(\X,\Y^a,\Q^a)$ by updating the model parameter $\Omega^a$ in an EM algorithm. Since $\Omega^a=\{\mu^a,\Sigma^a\}$ only relates to $\Phi(\x_i,Q^a_i|\Omega^a)$, which is a Gaussian distribution, we update $\Omega^a$ by
\begin{eqnarray}
\label{eqn:4}
&&\mu^a_\ell = \frac{1}{|N_\ell|}\sum_{i\in{N_\ell}}\x_i,\\
\label{eqn:5}
&&\Sigma^a_\ell = \frac{1}{|N_\ell|}\sum_{i\in{N_\ell}}(\x_i-\mu^a_\ell)\cdot(\x_i-\mu^a_\ell)^\mathsf{T},
\end{eqnarray}
where $N_\ell$ denotes the subset of face images in which $y^a_i =\ell$.


The optimization of the above two steps ends when the posterior $p(y^a_i|\mathbb{Y}^a_{\mathcal{N}_i},\x_i,Q^a_i,\Omega^a_t)$ converged. The output attribute $\Y^{a}$ is assigned with the final inferred $\Y^a_t$. Note that we use the original annotations. The optimization process is summarized in Algorithm~\ref{alg:label_propogation}.

\begin{figure}[b]
	\centering
	\includegraphics[width=1\linewidth]{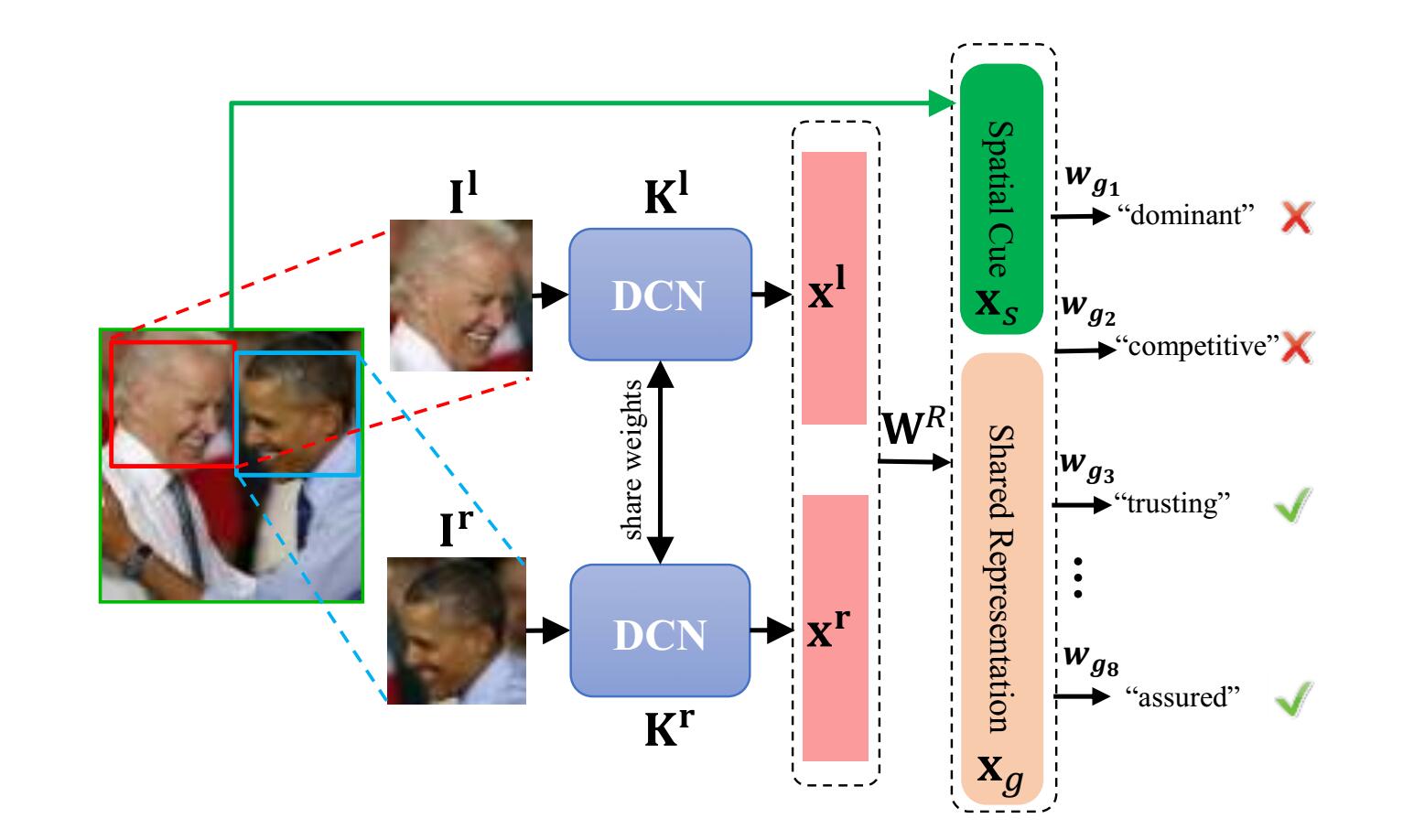}
	\caption{Overview of the network for interpersonal relation learning. The input is two face images and we extract the representation by two identical DCN, which is initialized by learning on multiple attribute datasets (see Sec.~\ref{sec:exp_representation}). Then we perform relation traits reasoning using face representation and additional spatial cues. The output is eight binary values that encode the different dimensions of relation traits.}
	\label{fig:social_network}
\end{figure}

\section{Interpersonal Relation Prediction from Face Images}
\label{sec:social_relation}

We have obtained a DCN that captures rich face representation through joint training with heterogeneous attribute sources. Next, we aim to jointly consider pairwise faces for interpersonal relation prediction.

We begin by arranging two identical DCNs obtained in Sec.~\ref{sec:exp_representation} in a Siamese-like architecture as shown in Fig.~\ref{fig:social_network}.
Using the interpersonal relation dataset introduced in Sec.~\ref{subsec:social_relation_dataset}, we train the new Siamese network end-to-end to map raw pixels of a pair of face images to relation traits.



As shown in Fig.~\ref{fig:social_network}, given an image with a detected pair of face, which is denoted as $\I^r$ and $\I^l$, we extract high-level features $\x^r$ and $\x^l$ using two DCNs respectively. These two DCNs have identical network structure as the one we use for expression recognition (see Sec.~\ref{sec:exp_representation}). Let $\K^r$ and $\K^l$ denote the network parameters. So we have $\forall\x^r,\x^l\in\mathbb{R}^{1024\times1}$. A weight matrix, $\W^R\in\mathbb{R}^{2048\times256}$, projects the concatenated feature vectors to a space of shared representation $\x_g \in\mathbb{R}^{256}$, which is utilized to predict a set of relation traits, $\g=\{g_i\}_{i=1}^8$, $\forall g_i\in\{0,1\}$. Each relation is modeled as a single binary classification task, parameterized by a weight vector, $\w_{g_i}\in\mathbb{R}^{256}$.

In addition to the face images, we incorporate some spatial cues to train the deep network as shown in Fig.~\ref{fig:social_network}. 
The spatial cues include:
\begin{enumerate}
\item Two faces' positions $\{x^l,y^l,w^l,h^l,x^r,y^r,w^r,h^r\}$, representing the $x$-,$y$-coordinates of the upper-left corner, width, and height of the bounding boxes; $w^l$ and $w^r$ are normalized by the image width. Similar for $h^{l}$ and $h^r$
\item The relative faces' positions: $\frac{x^l-x^r}{w^l},\frac{y^l-y^r}{h^l}$
\item The ratio between the faces' scales: $\frac{w^l}{w^r}$
\end{enumerate}
The above spatial cues are concatenated as a vector, $\x_s$, and combined with the shared representation $\x_g$ for learning relation traits.

Each binary variable $g_i$ can be predicted by linear regression,
\begin{equation}
g_i=\trans{\w}_{g_i}[\x_s;\x_g]+\epsilon,
\end{equation}
where $\epsilon$ is an additive error random variable, which is distributed following a standard logistic distribution, $\epsilon\sim Logistic(0,1)$. $[\cdot;\cdot]$ indicates the column-wise concatenation of two vectors.
Therefore, the probability of $g_i$ given $\x_g$ and $\x_s$ can be written as a sigmoid function, $p(g_i=1|\x_g,\x_s)=1/(1+\exp\{-\trans{\w}_{g_i}[\x_s;\x_g]\})$,
indicating that $p(g_i|\x_g,\x_s)$ is a Bernoulli distribution,
$p(g_i|\x_g,\x_s)=p(g_i=1|\x_g,\x_s)^{g_i}
\big(1-p(g_i=1|\x_g,\x_s)\big)^{1-g_i}$.

In addition, the probabilities of $\w_{g_i}$, $\W^R$, $\K^l$, and $\K^r$ can be modeled by the standard normal distributions. For example, suppose $\K$ contains $K$ filters, then $p(\K)=\prod_{j=1}^Kp(\kk_j)=\prod_{j=1}^K\mathcal{N}(\textbf{0},\mathcal{I})$, where $\textbf{0}$ and $\mathcal{I}$ are an all-zero vector and an identity matrix respectively, implying that the $K$ filters are independent. Similarly, we have $p(\w_{g_i})=\mathcal{N}(\textbf{0},\mathcal{I})$.
Furthermore, $\W^R$ can be initialized by a standard matrix normal distribution \cite{gupta1999matrix}, \ie, $p(\W^R)\propto\exp\{-\frac{1}{2}\mathrm{tr}(\W^R\trans{\W^R})\}$, where $\mathrm{tr}(\cdot)$ indicates the trace of a matrix.

Combining the above probabilistic definitions, the deep network is trained by maximizing a posterior probability,
\begin{equation}\label{eq:MAP}
\begin{split}
\arg&\max_{\Omega}~p(\{\w_{g_i}\}_{i=1}^8,\W,\K^l,\K^r|\g,\x_g,\x_s,\I^r,\I^l)\propto\\
&\Big(\prod_{i=1}^8 p(g_i|\x_g,\x_s)p(\w_{g_i})\Big)\Big(\prod_{j=1}^K p(\kk^l_j)p(\kk^r_j)\Big)p(\W^R),\\
&~~~~~~~~~~~~~~~~~~\mathrm{s.t.}~~~~~~\K^r=\K^l
\end{split}\vspace{-10pt}
\end{equation}
where $\Omega=\{\{\w_{g_i}\}_{i=1}^8,\W^R,\K^l,\K^r\}$ and the constraint means the filters are tied.
Note that $\x_g$ and $\x_s$ represent the hidden features and the spatial cues extracted from the left and right face images, respectively. Thus, the variable $g_i$ is independent with $\I^l$ and $\I^r$, given $\x_g$ and $\x_s$.

By taking the negative logarithm of Eqn.(\ref{eq:MAP}), it is equivalent to minimizing the following loss function
\begin{equation}\label{eq:E}
\begin{split}
&\arg\min_\Omega\sum_{i=1}^8\Big\{\trans{\w}_{g_i}\w_{g_i}-(1-g_i)\ln\big(1-p(g_i=1|\x_g,\x_s)\big)
-\\
&g_i\ln p(g_i=1|\x_g,\x_s)\Big\}+\sum_{j=1}^K(\trans{\kk^r_j}\kk^r_j
+\trans{\kk^l_j}\kk^l_j)+\mathrm{tr}(\W^R\trans{\W^R}),\\
&~~~~~~~~~~~~~~~~~~\mathrm{s.t.}~~~~~~\kk_j^r=\kk_j^l,~j=1...K
\end{split}
\end{equation}
where the second and the third terms correspond to the traditional cross-entropy loss, while the remaining terms indicate the weight decays~\cite{moody1995simple} of the parameters. Equation~(\ref{eq:E}) is defined over single training sample and is a highly nonlinear function because of the hidden features $\x_g$. Here we first initialize $\K^l$ and $\K^r$ by the representation we learn in Sec.~\ref{sec:exp_representation}. Then Eqn.~(\ref{eq:E}) is solved by stochastic gradient descent~\cite{krizhevsky2012imagenet}.

\section{Experiments}
\label{sec:experiments}

We divide our experiments into two subsections. Section~\ref{subsec:experiment_expression_attribute} examines the effectiveness of our base DCN on facial expression and attributes recognition. Section~\ref{subsec:experiment_interpersonal_relation} evaluates our full Siamese framework for interpersonal relation prediction.

\subsection{Facial Expression and Attributes Recognition}
\label{subsec:experiment_expression_attribute}

\noindent\textbf{Dataset}. 
We evaluated our base DCN on the combined dataset of AFLW, CelebA, and ExpW. From the total of 318,778 face images, we selected 5,400 images for testing and the remaining were reserved for training and validation.
The test images consisted of 3,000 CelebA, 1,000 AFLW, and 1,400 ExpW images.
We ensured that the ExpW test partition was balanced in their seven facial expression classes, \ie~all expression class had 200 samples. Note that this rule was not enforced in other attribute categories.

In addition to this combined dataset, we also evaluated our approach on the Static Facial Expressions in the Wild (SFEW) dataset~\cite{Dhall:2015:VIB:2818346.2829994} and CK+~\cite{5543262} datasets.

\begin{table*}[t]
    \newcommand{\tabincell}[2]{\begin{tabular}{@{}#1@{}}#2\end{tabular}}
    \caption{Balanced accuracies (\%) over different attributes.}
    \footnotesize
    \label{tab:attribute_acccuracy}
    \begin{center}
\setlength{\tabcolsep}{.1567em}
        \begin{tabular}{l|c|c|c|c|c|c|c|c|c|c|c|c|c|c|c|c|c|c|c|c|c|c|c|c|c}
            \hline
             \multirow{2}[10]{*}{Attributes}&\multirow{2}[33]{*}{\rotatebox{90}{$\qquad\qquad\;\;\;\;$average}}&Gender&\multicolumn{5}{c|}{Pose}&\multicolumn{10}{c|}{Expression}&\multicolumn{8}{c}{Age}\\\cline{2-26}
            &&\rotatebox{90}{gender}&\rotatebox{90}{left profile}&\rotatebox{90}{left}&\rotatebox{90}{frontal}&\rotatebox{90}{right}&
            \rotatebox{90}{right profile}&
            \rotatebox{90}{angry}&\rotatebox{90}{disgust}&\rotatebox{90}{fear}&
             \rotatebox{90}{happy}&\rotatebox{90}{sad}&\rotatebox{90}{surprise}&\rotatebox{90}{neutral}&\rotatebox{90}{smiling}&
            \rotatebox{90}{\parbox{1.1cm}{mouth \\ opened}}&
            \rotatebox{90}{\parbox{1.1cm}{narrow \\ eyes}}&
            \rotatebox{90}{young}&\rotatebox{90}{goatee}&
            \rotatebox{90}{no beard}&\rotatebox{90}{sideburns}&
            \rotatebox{90}{\parbox{1.5cm}{5 o'clock \\ shadow}}&
            \rotatebox{90}{gray hair}&
            \rotatebox{90}{bald}&
            \rotatebox{90}{mustache}\\
            \hline\hline
            HOG+SVM&71.0&83.2&73.8&65.7&88.3&60.3&70.1&54.3&54.8&56.2&71.3&58.4&61.2&68.4&84.5&79.7&56.3&72.9&75.6&88.4&75.8&72.4&85.9&75.4&70.4\\
            Baseline DCN&76.1&96.5&75.0&56.3&87.3&51.8&74.2&63.5&50.0&50.0&81.9&64.0&71.0&75.0&93.0&94.2&63.3&\textbf{84.4}&84.8&92.8&88.6&82.8&87.7&86.1&73.4\\
            DCN+AP&\textbf{80.9}&\textbf{97.0}&\textbf{78.4}&\textbf{67.3}&\textbf{90.0}&\textbf{62.1}&\textbf{77.9}&\textbf{72.1}&\textbf{56.5}&\textbf{58.7}&\textbf{83.8}&\textbf{69.1}&\textbf{74.2}&\textbf{76.0}&\textbf{93.3}&\textbf{94.5}&\textbf{73.5}&83.5&\textbf{90.1}&\textbf{92.5}&\textbf{92.5}&\textbf{88.3}&\textbf{92.2}&\textbf{93.0}&\textbf{83.9}\\
            \hline
        \end{tabular}
    \end{center}
\end{table*}

\vspace{0.1cm}
\noindent\textbf{Evaluation metric}.
To account for the imbalanced positive and negative attribute samples, a balanced accuracy is adopted as the evaluation metric:
\begin{equation}\label{eq:evaluation}
accuracy=0.5\times(n_p/N_p+n_n/N_n),
\end{equation}
where $N_p$ and $N_n$ are the numbers of positive and negative samples, while $n_p$ and $n_n$ are the numbers of true positive and true negative.

\vspace{0.2cm}
\noindent\textbf{Implementation.} We implemented the proposed deep model with MXNet~\cite{mxnet} library. 
Data augmentation by random translation and mirroring were introduced in the training process. The mini-batch size was fixed to 32, and the learning rate was 0.001 with a momentum rate of 0.9. Following Algorithm~\ref{alg:Framework}, the first initialization stage took 30 epochs to converge, while the second stage on attribute propagation consumed another 10 epochs (\ie, $M=10$).

        

\vspace{0.2cm}
\noindent\textbf{Results on the combined AFLW, CelebA, and ExpW.} 
We trained two variants of our DCN using the combined dataset: 
\begin{enumerate}
\item Baseline DCN - it is trained without both attribute propagation.
\item DCN+AP - it is trained with attribute propagation (\ie~full model).
\end{enumerate}

For completeness, we additionally trained a baseline classifier by extracting HOG features from the given face images, and we used a linear support vector machine (SVM) to train a binary classifier (\ie, HOG+SVM) for each attribute. In the SVM learning process, we adjusted the weight of each class as inversely proportional to the class frequency in the training data. This helped in mitigating the imbalanced class issue.

The balanced accuracy of each method is reported in Table~\ref{tab:attribute_acccuracy}. It is observed that in general, attribute propagation helps, especially on attributes with rare positive samples such as ``narrow eyes'' and ``goatee''. We conjecture that attribute propagation allows the proposed model to effectively leverage samples from multiple datasets, which are not annotated initially.


To further compare with existing attribute recognition methods, we follow the training and testing splits of CelebA \cite{ziwei} (as for AFLW and ExpW, we use the same training data as the previous experiments). The performance is summarized in Table~\ref{tab:celeba_acc}. Note that we follow the convention of~\cite{ziwei}, and use the overall classification accuracy instead of the balanced accuracy as Eqn.~(\ref{eq:evaluation}). We can observe that by fusing multiple datasets, our proposed method achieves superior performance compared to state-of-the-art methods. 

\begin{figure}
	\centering
	\includegraphics[width=1\linewidth]{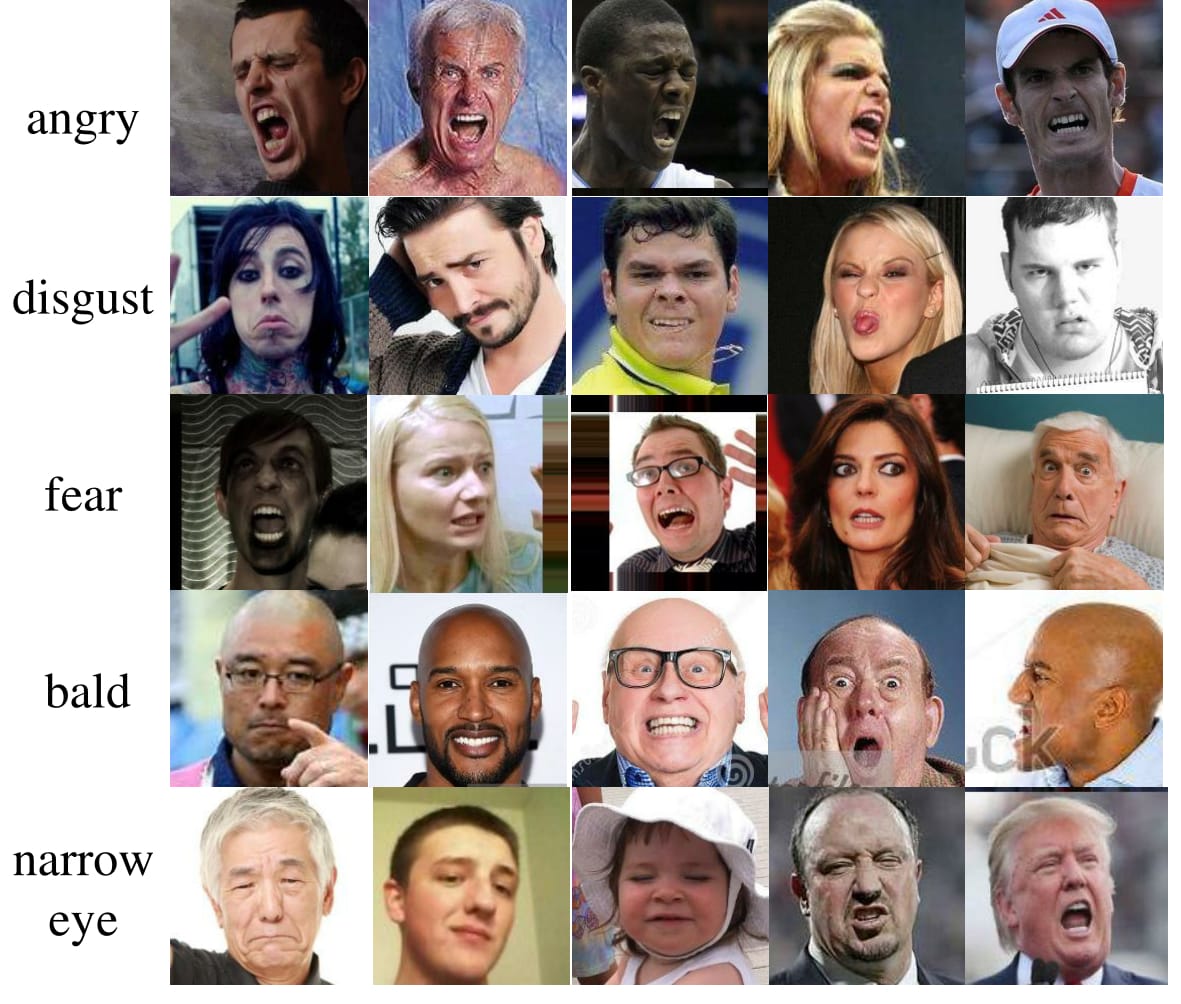}
	\caption{Examples of automatically annotated positive attribute examples via the proposed attribute propagation (discussed in Sec.~\ref{subsec:attribute_propagation}).}
	\label{fig:lp}
\end{figure}

In Table~\ref{tab:m_acc}, we show the average balanced accuracy over different iterations of the alternating attribute propagation and representation learning process (see Sec.~\ref{subsec:attribute_propagation}). 
The gradually improved accuracy over iterations demonstrates that the alternating optimization process is beneficial. Figure~\ref{fig:lp} shows a few initially unlabeled positive attribute samples that are automatically annotated via attribute propagation.
It is worth pointing out that many of this unlabeled samples are challenging in terms of their unconstrained poses and expressions.


\begin{table}[t]
	\centering
	\caption{Attribute Recognition Accuracy on CelebA.}
	\vspace{0.1cm}
	\label{tab:celeba_acc}
	\setlength{\tabcolsep}{.2367em}
	\begin{tabular}{c|c|c|c|c|c|c|c|c|c|c|c|c}
		
		\hline  Method&\rotatebox{90}{smiling}&
			\rotatebox{90}{\parbox{1.1cm}{mouth \\ opened}}&
			\rotatebox{90}{\parbox{1.1cm}{narrow \\ eyes}}&
			\rotatebox{90}{young}&\rotatebox{90}{goatee}&
			\rotatebox{90}{no beard}&\rotatebox{90}{sideburns}&
			\rotatebox{90}{\parbox{1.5cm}{5 o'clock \\ shadow}}&
			\rotatebox{90}{gray hair}&
			\rotatebox{90}{bald}&
			\rotatebox{90}{mustache}&\rotatebox{90}{gender}\\
		\hline
		FaceTracer~\cite{kumar2008facetracer}&89&87&82&80&93&90&94&85&90&89&91&91\\
		PANDA-w~\cite{zhang2014panda}&89&82&79&77&86&87&90&82&88&92&83&93\\
		PANDA-l~\cite{zhang2014panda}&92&93&84&84&93&93&93&88&94&96&93&97\\
		Liu~\etal~\cite{ziwei}&92&92&81&87&95&95&96&91&97&98&95&\textbf{98}\\
		MCNN-AUX~\cite{aaaiattribute}&93&94&87&88&97&96&98&95&98&\textbf{99}&97&\textbf{98}\\
		ours&                          \textbf{94}&\textbf{95}&\textbf{89}&\textbf{91}&\textbf{98}&\textbf{97}&\textbf{98}&\textbf{96}&\textbf{99}&\textbf{99}&\textbf{98}&\textbf{98}\\
		\hline
	\end{tabular}
\end{table}

\begin{table}[t]
	\centering
	\caption{Average balanced accuracies (\%) over different iterations of the alternating attribute propagation and representation learning process.}
	\vspace{0.1cm}
	\label{tab:m_acc}
	\begin{tabular}{c|c|c|c|c|c|c}
		
		\hline  Iteration &M=1&M=3&M=5&M=7&M=9&M=10\\
		\hline
		Accuracy&78.4&79.2&79.3&79.8&80.8&80.9\\
		\hline
	\end{tabular}
\end{table}

\vspace{0.2cm}
\noindent\textbf{Expression Recognition on SFEW~\cite{Dhall:2015:VIB:2818346.2829994}.}
To demonstrate the effectiveness of the proposed DCN for facial expression recognition, we evaluated its performance on the challenging Static Facial Expressions in the Wild (SFEW) 2.0 dataset~\cite{Dhall:2015:VIB:2818346.2829994}. 
The dataset is a static subset of Acted Facial Expressions in the Wild (AFEW) dataset~\cite{Dhall:2015:VIB:2818346.2829994}, which captures natural and versatile expressions from movies.
Since the label for the test set is not publicly available, we follow the training/validation splits of the released dataset, we evaluated two variants of our method: 1) Our trained DCN+AP without fine-tuning on SFEW training partition, and 2) Our trained full model DCN+AP with fine-tuning on SFEW training partition. 
Our model treats each expression as a binary attribute, the expression with the highest predicted probability is selected as the classification result. 

We compared our method with the following approaches: 
\begin{enumerate}
\item PHOG+LPQ~\cite{Dhall:2015:VIB:2818346.2829994} - the Pyramid of Histogram of Gradients (PHOG) and Local Phase Quantization (LPQ)~\cite{dhall2011emotion} are computed and concatenated to form the feature of a face, and a non-linear SVM is used for expression classification.

\item MBP~\cite{Levi:2015:ERW:2818346.2830587} - expression recognition with Mapped Binary Patterns (MBP), which is proposed in~\cite{Levi:2015:ERW:2818346.2830587}.

\item AU-Aware Features~\cite{Yao:2015:CAF:2818346.2830585} - expression recognition by exploiting facial action-unit aware features.

\item Microsoft Emotion API~\cite{Microsoft} - emotion API of Microsoft cognitive services. Since it is a commercial API, we use the service directly without fine-tuning in on the SFEW training partition.

\item DCN of~\cite{Ng:2015:DLE:2818346.2830593} - AlexNet~\cite{krizhevsky2012imagenet} pretrained with ImageNet~\cite{ILSVRC15} and FER~\cite{Goodfeli-et-al-2013} datasets, and finetune on the SFEW training dataset.

\item DCN of~\cite{yu2015image} - A customized DCN (five convolutional layers and two fully connected layers) pretrained with FER~\cite{Goodfeli-et-al-2013} dataset, and fine-tune on the SFEW training dataset.

\end{enumerate}

\begin{figure}
	\centering
	\includegraphics[width=1\linewidth]{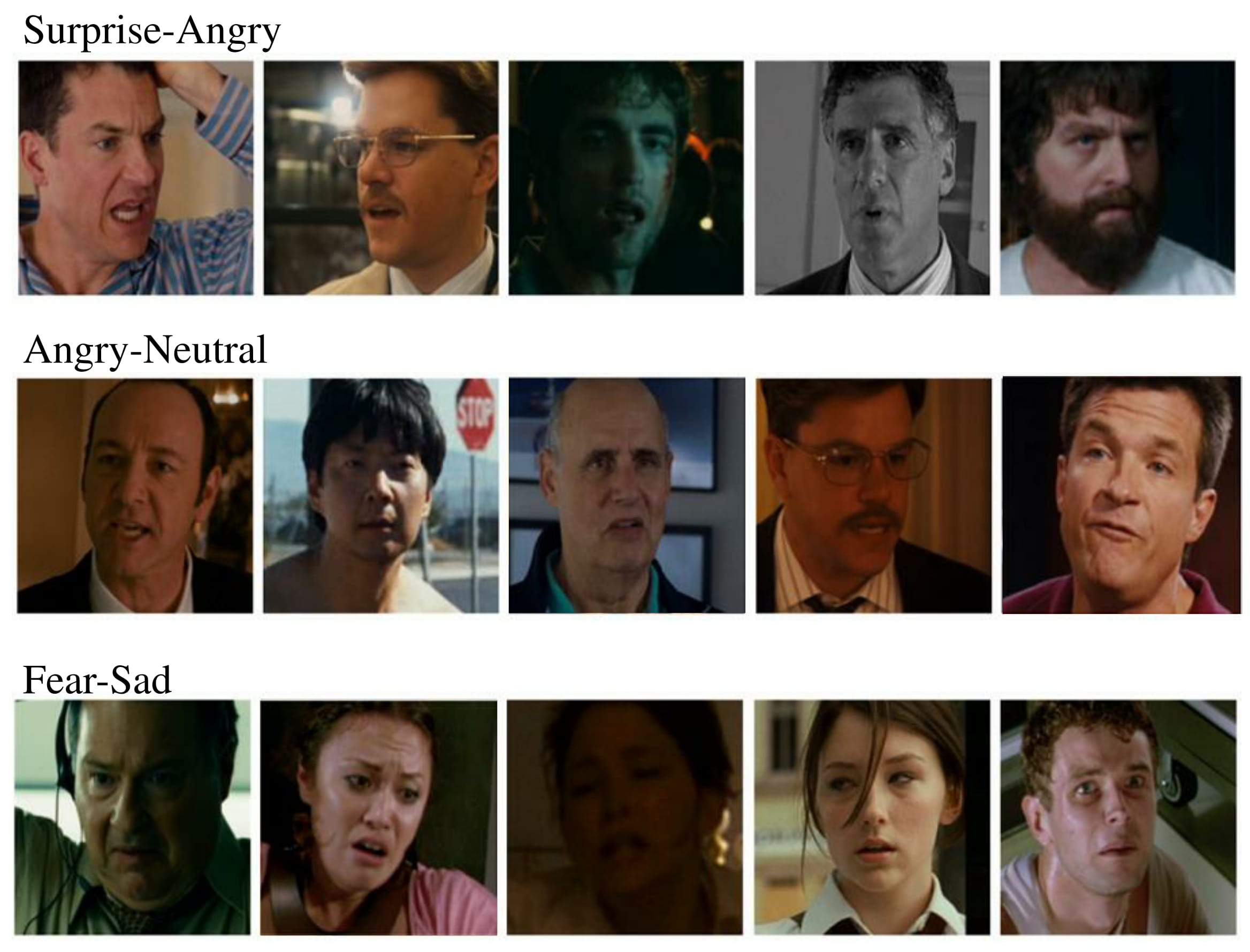}
	\caption{Example of failure cases of our approach (DCN+AP) on the SFEW validation set. The text above each row denotes the ground truth and predicted result, \eg, ``Surprise-Angry'' means the surprise expression is misclassified as angry. Most failures were caused by ambiguity in facial expressions.}
	\label{fig:sfew_failure}
\end{figure}

Table~\ref{tab:sfew_acc} summarizes the performances of various approaches evaluated on the SFEW dataset.
Following the convention of current studies~\cite{yu2015image,liu2015inspired,khorrami2015deep,liu2013facial}, we use the overall classification accuracy instead of the balanced accuracy as Eqn.~(\ref{eq:evaluation}).
Our approach, with and without fine-tuning on SFEW training partition, outperforms state-of-the-art methods. Again, it is observed that our model is benefited from alternating optimization with attribute propagation. Figure~\ref{fig:sfew_failure} shows some failure cases. Most errors were caused by ambiguous cases.

\begin{table}[t]
	\centering
	\caption{Accuracies on the validation set of SFEW dataset~\cite{Dhall:2015:VIB:2818346.2829994}.}
	\vspace{0.1cm}
	\label{tab:sfew_acc}
	\setlength{\tabcolsep}{.10667em}
	\begin{threeparttable}
	\begin{tabular}{c|c|c}
		
		\hline  Method&\parbox[t]{3cm}{Training/Fine-tunning\\on SFEW}  &Accuracy\\
		\hline
		PHOG+LPQ~\cite{Dhall:2015:VIB:2818346.2829994}&yes&35.93\%\\
		MBP~\cite{Levi:2015:ERW:2818346.2830587}&yes&41.92\%\\
		AU-Aware features~\cite{Yao:2015:CAF:2818346.2830585}&yes&44.04\%\\
		Microsoft Emotion API~\cite{Microsoft}&no&47.71\%\\
		DCN of~\cite{Ng:2015:DLE:2818346.2830593}&yes&48.50\%\\
		Single DCN of~\cite{yu2015image} &yes&52.29\%\\
		Ensemble DCNs of~\cite{yu2015image}\footnotemark[1]&yes&55.96\%\\
		\hline
		\hline
		Our Baseline DCN&no&45.51\%\\
		Our DCN+AP&no&49.77\%\\
		Our Baseline DCN&yes&52.06\%\\
		Our DCN+AP&yes&55.27\%\\
		\hline
	\end{tabular}
	 \begin{tablenotes}
        \footnotesize
        \item[1] This result is obtained from an ensemble of five DCNs.
      \end{tablenotes}
    \end{threeparttable}
\end{table}

\vspace{0.2cm}
\noindent\textbf{Expression Recognition on CK+~\cite{5543262}.} 
For completeness, we also evaluated our method on CK+~\cite{5543262} since it is a classic dataset for expression recognition. CK+ contains 327 image sequences where each sequence presents a face with gradual expression evolvement from a neutral to a peak facial expression. 
Each sequence is annotated with one of the six prototypical expressions, \ie,~angry, happy, surprise, sad, disgust, fear, or a non-standard expression (\ie~contempt). 
Following the widely used evaluation protocol~\cite{liu2013facial,khorrami2015deep,zhao2016peak}, we selected the last three frames of each sequence for training/testing purpose. The first frame of each sequence was regarded as the ``neutral'' expression. 
Consequently, we obtained 1,308 images for our 10-fold cross-validation. The face identity in each fold was remained exclusive. As in the SFEW experiments, we fine-tuned our trained DCN+AP on the training samples of each fold. 

Table~\ref{tab:ck+_acc} presents the comparative results of our method and other state-of-the-arts. To be consistent with other methods, the averaged accuracy of the six basic expressions are reported. Similar to our approach, BDBN~\cite{liu2013facial}, PPDN~\cite{zhao2016peak}, and Zero-bias CNN~\cite{khorrami2015deep} also adopted different kinds of deep networks. Our approach still achieves better result although the performance on CK+ is nearly saturated.

\begin{table}[t]
	\centering
	\caption{Accuracies on the CK+ dataset~\cite{5543262} with six prototypical facial expressions.}
	\vspace{0.1cm}
	\label{tab:ck+_acc}
	\setlength{\tabcolsep}{.10667em}
	\begin{tabular}{c|c}
		
		\hline  Method&Accuracy\\
		\hline
		CSPL~\cite{6247974}&89.9\%\\
		LBPSVM\cite{Shan:2009:FER:1523527.1523949}&95.1\%\\
		BDBN~\cite{liu2013facial}&96.7\%\\
		PPDN~\cite{zhao2016peak}&97.3\%\\
		Zero-bias CNN~\cite{khorrami2015deep}&98.3\%\\
		Our Method&98.9\%\\
		\hline
	\end{tabular}
\end{table}

%

\begin{figure*}[t]
	\centering
	\includegraphics[width=1\textwidth]{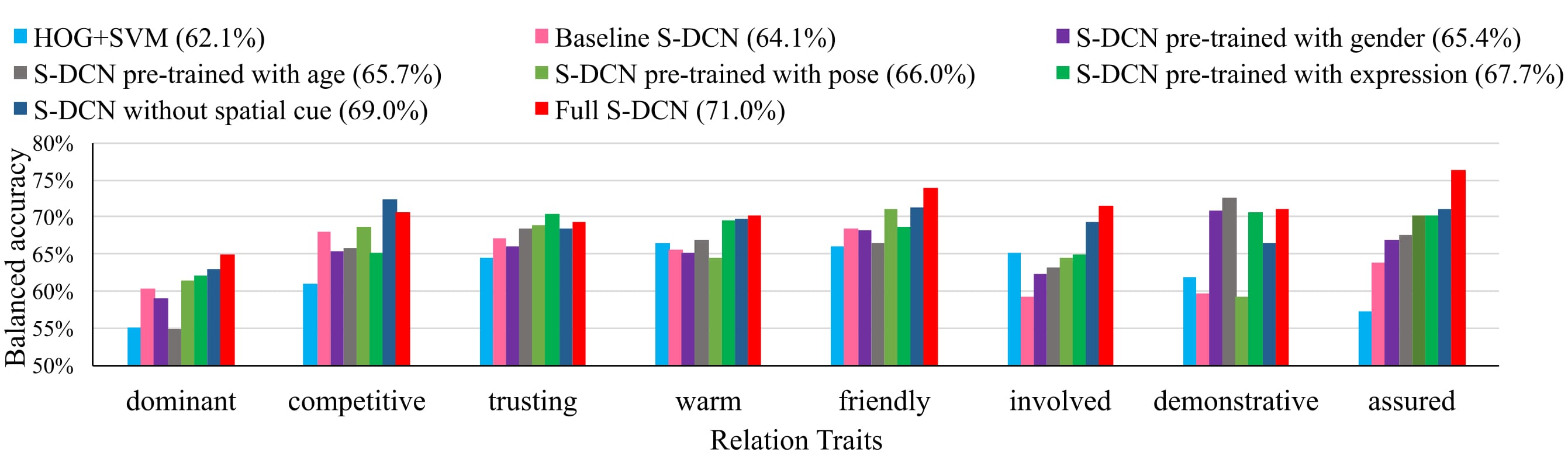}
	\caption{Relation traits prediction performance. The number in the legend indicates the average accuracy of the according method across all the relation traits.}
	\label{fig:performance}
\end{figure*}

\subsection{Interpersonal Relation Prediction}
\label{subsec:experiment_interpersonal_relation}

\begin{table}[t]
	\caption{Statistics of the interpersonal relation dataset.}
	\footnotesize
	\label{tab:relation_datasets}
	\begin{center}
		\setlength{\tabcolsep}{.58667em}
		\begin{tabular}{l|c|c|c|c}
			\hline
			\multirow{2}{*}{Relation trait}&\multicolumn{2}{c|}{training}&\multicolumn{2}{c}{testing}\\\cline{2-5}
			&\#positive&\#negative&\#positive&\#negative\\
			\hline \hline
			dominant&418&6808&112&678\\
			
			competitive&344&6882&70&720\\
			
			trusting&6261&965&606&184\\
			
			warm&6176&1050&615&175\\
			
			friendly&6733&493&728&62\\
			
			involved&6360&866&686&104\\
			
			demonstrative&6494&732&689&101\\
			
			assured&6538&688&673&117\\
			\hline
		\end{tabular}
	\end{center}
\end{table}

\noindent
\textbf{Dataset}. The evaluation of interpersonal relation learning was performed on the dataset described in Sec.~\ref{subsec:social_relation_dataset}. We divided the dataset into training and test partitions of 7,226 and 790 images, respectively. The face pairs in these two partitions were mutually exclusive, containing no overlapped identities. Table~\ref{tab:relation_datasets} presents the statistics of this dataset. 

\vspace{0.1cm}
\noindent
\textbf{Evaluation metric}. We adopt the same balanced accuracy in Eqn.~(\ref{eq:evaluation}).

\vspace{0.1cm}
\noindent
\textbf{Baselines}. 
As discussed in Sec.~\ref{sec:social_relation}, our full model combines the two DCNs pre-trained for expression and attribute recognition in a Siamese-like architecture, as shown in Fig.~\ref{fig:social_network}. We call this model as ``S-DCN''. 

We evaluated several variants of this network.
\begin{enumerate}
\item Baseline S-DCN - We trained a model similar to S-DCN in Fig.~\ref{fig:social_network}, but without using the DCN pre-trained for expression and attribute recognition. Instead, the parameters of the two DCNs were randomly initialized.

\item S-DCN with its DCN pre-trained with selected attributes - To examine the influences of different attribute groups, we pre-trained four DCN variants using only one group of attribute (\ie, expression, age, gender, and pose), respectively.

\item S-DCN without spatial cue - We trained a S-DCN with DCN pre-trained with all the attributes but the spatial cue (discussed in Sec.~\ref{sec:social_relation}) was not used.

\item Full S-DCN - We trained a S-DCN with DCN pre-trained with all the attributes and used the spatial cue as discussed in Sec.~\ref{sec:social_relation}. 
\end{enumerate}

In addition, we established a baseline ``HOG+SVM'' - we extracted the HOG features from the given face images. The features from two faces were then concatenated and a linear support vector machine (SVM) was employed to train a binary classifier for each relation trait.

\vspace{0.1cm}
\noindent
\textbf{Results}. 
Figure~\ref{fig:performance} shows the accuracies of different variants.
All variants of the proposed S-DCN outperform the baseline HOG+SVM.
We observe that the cross-dataset expression and attribute pre-training is beneficial since pre-training with any of the attribute groups improves the overall performance.
In particular, pre-training with expression attributes outperforms other groups of attributes (improving from 64.1\% to 67.7\%). This is not surprising since interpersonal relation is largely manifested from expression.
The pose attributes come next in terms of influence to relation prediction.
The result is also expected since when people are in a close or friendly relation, they tend to look at the same direction or face each other.

Finally, the spatial cue is shown to be useful for relation prediction. However, we also observe that not every trait is improved by the spatial cue and some are degraded. This is because currently we simply use the face scale and location directly, of which the distribution is inconsistent in images from different sources. For example, some close-shot photographs may be used to show competing people and their expression in detail, while in some movies, competing people may stand far away from each other. As for the relation traits, ``dominant'' is the most difficult trait to predict as it needs to be determined by more complicated factors, such as one's social role and the environmental context.

\begin{figure}
    \centering
    \includegraphics[width=0.5\textwidth]{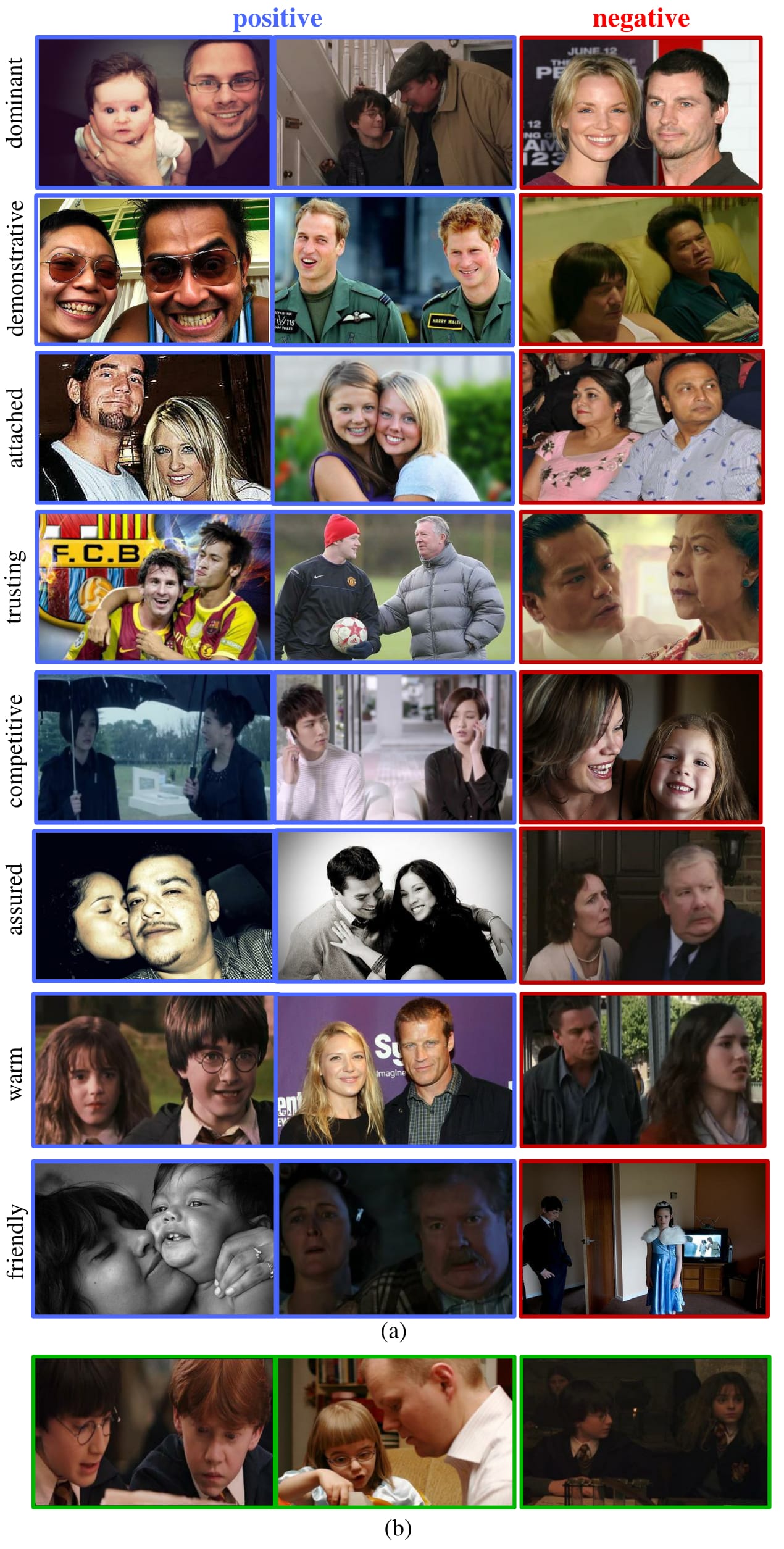}
    \caption{(a) Correct positive and negative prediction results on different relation traits. (b) False positives on ``competitive'', ``assured'' and ``demonstrative'' relation traits (from left to right).}
    \label{fig:samples2}
\end{figure}

\begin{table}
    \centering
    \caption{Balanced accuracies (\%) on the movie testing subset.}
    \label{tab:movie_subset}
    \setlength{\tabcolsep}{.30667em}
    \begin{tabular}{c|c}
        
        \hline  Method&Balanced Accuracy\\
        \hline
        \hline  HOG+SVM& 59.22\% \\
        \hline  Baseline S-DCN&62.42\%\\
        \hline  S-DCN (DCN pre-trained with gender)&63.10\%\\
        \hline  S-DCN (DCN pre-trained with age)&64.67\%\\
        \hline  S-DCN (DCN pre-trained with pose)&62.83\%\\
        \hline  S-DCN (DCN pre-trained with expression)&65.36\%\\
        \hline  S-DCN without spatial cue&68.17\%\\
        \hline  Full S-DCN&70.20\%\\

        \hline
    \end{tabular}
\end{table}

To factor out any potential subjective judgement arisen from the data annotation process, we evaluated S-DCN on a subset of 522 movie frames extracted from the test data. 
This subset is more `objective' since annotators were provided with richer auxiliary cues for relation annotation. 
Table~\ref{tab:movie_subset} shows the average balanced accuracy on the eight relation traits of the baseline and the variants of the proposed S-DCN. The results further suggest the reliability of the proposed approach.

\begin{figure}
    \centering
    \includegraphics[width=0.45\textwidth]{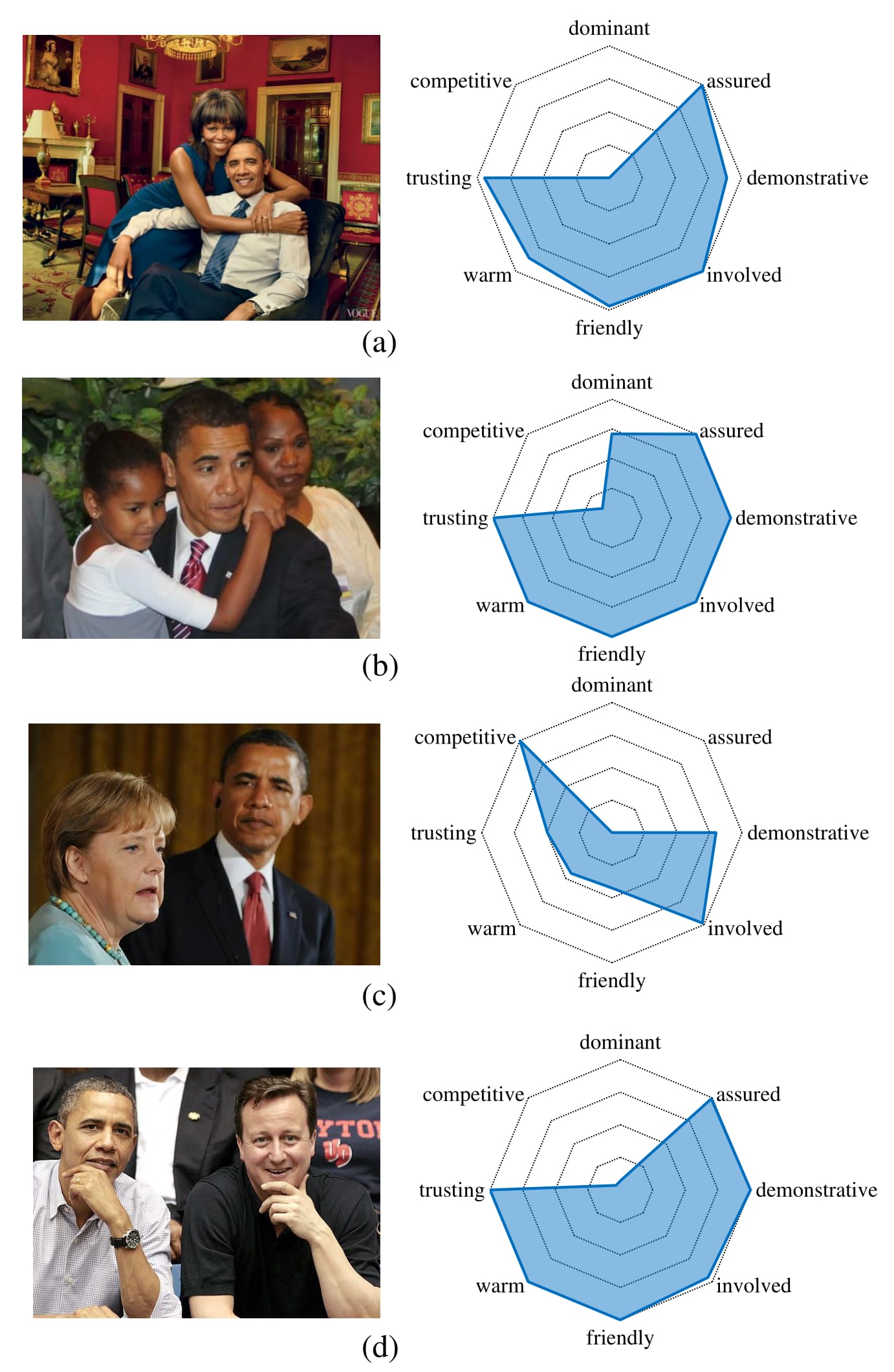}
    \caption{The relation traits predicted by our full model with spatial cue (Full S-DCN). The polar graph beside each image indicates the tendency for each trait to be positive.}
    \label{fig:samples}
\end{figure}

\begin{figure*}
    \centering
    \includegraphics[width=0.75\textwidth]{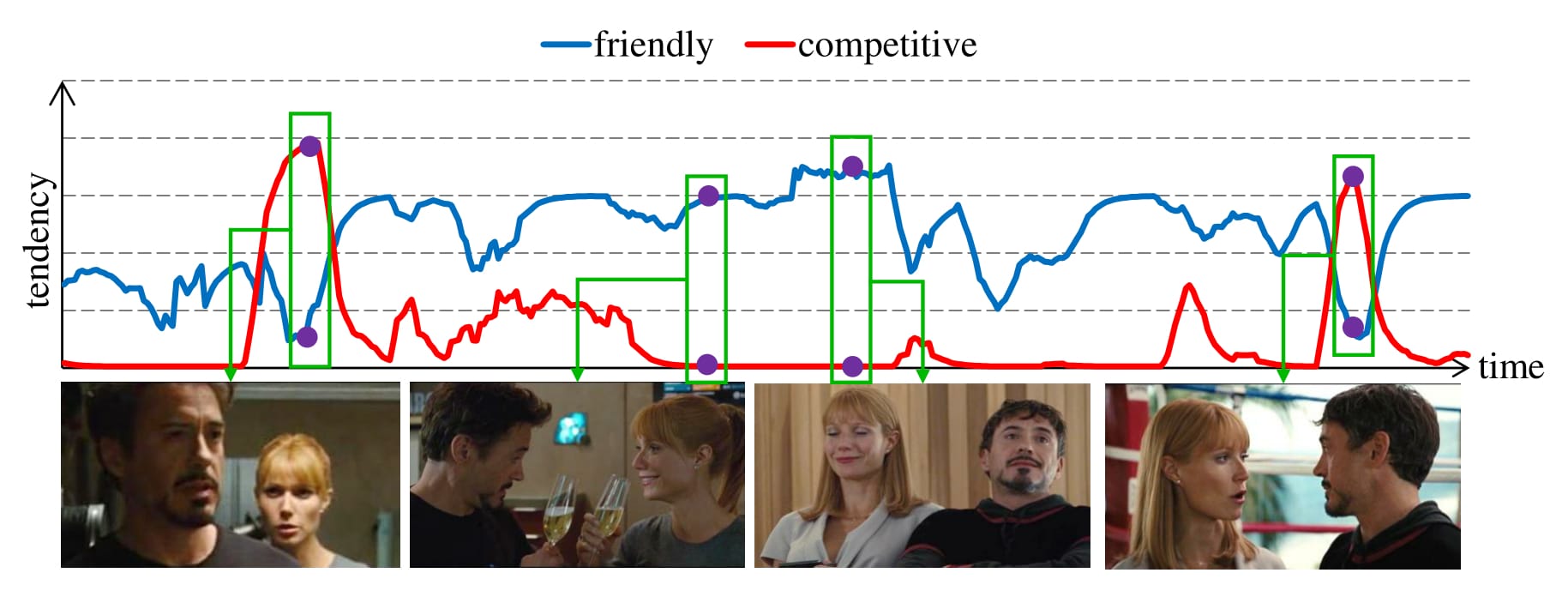}
    \caption{Prediction for relation traits of ``friendly'' and ``competitive'' for the movie \emph{Iron Man}. The probability indicates the tendency for the trait to be positive. It shows that the proposed approach can capture the friendly talking scene and the moment of conflict.}
    \label{fig:vis_video}
\end{figure*}

Some positive and negative predictions on different relation traits are shown in Fig.~\ref{fig:samples2}(a). It can be observed that the proposed approach is capable of handling images in different scenes and faces with large expression variations. We show some false positives in Fig.~\ref{fig:samples2}(b), which are partly caused by the lack of context. For example, in the first image of Fig.~\ref{fig:samples2}(b), the two characters were having a serious conversation. The algorithm had no access to the context that they were reading a book and thus guessed that they were competing.
Our method also failed given faces with a large degree of occlusions.

More qualitative results are presented in Fig.~\ref{fig:samples}. Positive relation traits, such as ``trusting'', ``warm'', ``friendly'' are inferred between the US President \textit{Barack Obama} and his family members. Interestingly, ``dominant'' trait is predicted between him and his daughter (Fig.~\ref{fig:samples} (b)). Fig.~\ref{fig:samples}(c) includes the image for \emph{Angela Merkel}, Chancellor of Germany, which is usually used in the news articles on US spying scandal, showing a low tendency on the ``trusting'' trait, while a high tendency on the ``competitive'' trait. This relation is quite different from that of Fig.~\ref{fig:samples}(d), where \textit{Obama} and the British Prime Minister \textit{David Cameron} were watching a basketball game.


We show an example of application of using our method to automatically profile the relations among the characters in a movie. We chose the movie \emph{Iron Man} and focused on different interaction patterns, such as conversation and conflict, of the main roles ``\emph{Tony Stark}'' and ``\emph{Pepper Potts}''. Firstly, we applied a face detector to the movie and selected those frames that captured the two roles. Then, we applied our approach on each frame to infer their relation traits. The predicted probabilities were averaged across 5 neighboring frames to obtain a smooth profile. Figure~\ref{fig:vis_video} shows a video segment with the traits of ``friendly'' and ``competitive''. Our method accurately captures the friendly talking scene and the moment when Tony and Pepper were in a conflict, where the ``competitive'' trait is assigned with a high probability while the ``friendly'' trait is low.

\section{Conclusion}

In this work, we studied a new challenging problem of predicting interpersonal relation from face images. 
We decomposed our solution into two steps. We began with training a reliable deep convolutional network for recognizing facial expression and rich attributes (gender, age, and pose) from single face images.
We addressed the problem of learning from heterogeneous data sources with potentially missing attribute labels. This was achieved through a novel approach that leverages the inherent correspondences among heterogeneous sources by attribute propagation in a graphical model.
Initialized by the deep convolutional network learned in the first step, a Siamese-like framework is proposed to learn an end-to-end mapping from raw pixels of a pair of face images to relation traits. 
Extensive experiments demonstrate the effectiveness of the proposed methods on facial expression recognition and interpersonal relation prediction.
Future work will combine the face-based relation traits with body-driven immediacy cues~\cite{chu2015multi} for more accurate interpersonal relation prediction.  

\begin{acknowledgements}
This work is supported by SenseTime Group Limited and the General Research Fund sponsored by the Research Grants Council of the Hong Kong SAR (CUHK 14241716, 14224316. 14209217).
\end{acknowledgements}

\bibliographystyle{spmpsci}      
\bibliography{long,zp_bib_v2.1}   

\end{document}